\documentclass[sigconf]{acmart}
\usepackage{graphicx}
\usepackage{array}
\usepackage{longtable}
\usepackage{multirow}
\usepackage{tabularx}
\usepackage{bbm}
\usepackage{xcolor}
\usepackage{hyperref}

\definecolor{linkblue}{RGB}{0,70,140}

\hypersetup{
    colorlinks=true,
    linkcolor=linkblue,
    citecolor=linkblue,
    urlcolor=linkblue
}

\setcopyright{acmlicensed}
\copyrightyear{2026}
\acmYear{2026}
\setcopyright{cc}
\setcctype{by}
\acmConference[SIGSPATIAL '26]{The 34th ACM International Conference on Advances in Geographic Information Systems}{November 03--06, 2026}{Riverside, CA, USA}
\acmBooktitle{The 34th ACM International Conference on Advances in Geographic Information Systems (SIGSPATIAL '26), November 03--06, 2026, Riverside, CA, USA}
\acmDOI{10.1145/3841645.3842981}
\acmISBN{979-8-4007-2950-8/2026/11}

\newcommand{\method}{SPECTRA}
\newcommand{\bre}{BRE}
\newcommand{\starplanner}{ST-LoRA}
\newcommand{\miou}{mIoU}
\newcommand{\bandsel}{band-selection}
\newcommand{\R}{\mathbb{R}}

\begin{document}

\title[\method{}]{\method{}: Band-Routed Embedding and Stage-Wise LoRA for Cross-Sensor Fine-Tuning of Geospatial Foundation Models}

\author{Xingyan Li}
\email{xingyanli@umbc.edu}
\affiliation{%
  \institution{University of Maryland, Baltimore County}
  \city{Baltimore}
  \state{Maryland}
  \country{USA}}
% \email{email@example.com}
\renewcommand{\shortauthors}{Author et al.}

\author{Jordan A. Caraballo-Vega}
\email{jordan.a.caraballo-vega@nasa.gov}
\affiliation{%
  \institution{NASA Goddard Space Flight Center}
  \city{Greenbelt}
  \state{Maryland}
  \country{USA}}
% \email{email@example.com}
\renewcommand{\shortauthors}{Author et al.}

\author{Jie	Gong}
\email{jie.gong@nasa.gov}
\affiliation{%
  \institution{NASA Goddard Space Flight Center}
  \city{Greenbelt}
  \state{Maryland}
  \country{USA}}
% \email{email@example.com}
\renewcommand{\shortauthors}{Author et al.}

\author{Mark L. Carroll}
\email{mark.carroll@nasa.gov}
\affiliation{%
  \institution{NASA Goddard Space Flight Center}
  \city{Greenbelt}
  \state{Maryland}
  \country{USA}}
% \email{email@example.com}
\renewcommand{\shortauthors}{Author et al.}

\author{Jianwu Wang}
\email{jianwu@umbc.edu}
\affiliation{%
  \institution{University of Maryland, Baltimore County}
  \city{Baltimore}
  \state{Maryland}
  \country{USA}}
% \email{email@example.com}
\renewcommand{\shortauthors}{Li et al.}

\begin{abstract}

Geospatial foundation models (GeoFMs), pretrained on large-scale geospatial data such as Earth observation (EO), climate, and weather data, have shown promising performance when fine-tuned on diverse downstream tasks. However, there are two challenges in adapting EO-pretrained GeoFMs to practical fine-tuning datasets (downstream datasets). The first challenge is how to handle spectral mismatch: pretrained patch embeddings expect a fixed set of input bands, whereas downstream sensors may provide different channels. The second challenge is how to reduce fine-tuning cost and make it efficient. While existing work has made efforts on these challenges individually, methods that tackle both jointly remain underexplored. We propose SPECTRA, a parameter-efficient fine-tuning framework that addresses both spectral mismatch and adaptation cost. To handle spectral mismatch, SPECTRA introduces Band-Routed Embedding (BRE), which maps all available downstream bands into the band space expected by the pretrained GeoFM. By using BRE, all available bands in the downstream dataset are utilized to improve the selected-band input without changing the pretrained patch embedding interface. To reduce adaptation cost, SPECTRA further introduces a Stage-wise Transferability-aware LoRA (ST-LoRA) fine-tuning. ST-LoRA estimates stage-wise transferability before fine-tuning and assigns stage-specific LoRA ranks to reduce trainable parameters. Across three EO-pretrained GeoFMs and four downstream segmentation datasets, experiments show that BRE improves performance by utilizing all spectral bands, while ST-LoRA reduces trainable parameters compared with full fine-tuning and standard LoRA. Code is available at \href{https://github.com/big-data-lab-umbc/SPECTRA}{\texttt{github.com/big-data-lab-umbc/SPECTRA}}.

\end{abstract}

\begin{CCSXML}
<ccs2012>
 <concept>
  <concept_id>10010147.10010257.10010258.10010262.10010277</concept_id>
  <concept_desc>Computing methodologies~Transfer learning</concept_desc>
  <concept_significance>500</concept_significance>
 </concept>
 <concept>
  <concept_id>10010147.10010178.10010224.10010245.10010247</concept_id>
  <concept_desc>Computing methodologies~Image segmentation</concept_desc>
  <concept_significance>300</concept_significance>
 </concept>
 <concept>
  <concept_id>10010405.10010432.10010437</concept_id>
  <concept_desc>Applied computing~Earth and atmospheric sciences</concept_desc>
  <concept_significance>300</concept_significance>
 </concept>
</ccs2012>
\end{CCSXML}

\ccsdesc[500]{Computing methodologies~Transfer learning}
\ccsdesc[300]{Computing methodologies~Image segmentation}
\ccsdesc[300]{Applied computing~Earth and atmospheric sciences}

\keywords{Geospatial foundation models, Cross-sensor adaptation, Parameter-efficient fine-tuning, LoRA, Multispectral imagery}

\maketitle

\section{Introduction}

Geospatial foundation models (GeoFMs) are increasingly used as reusable backbones for Earth-observation (EO) tasks. Models such as Prithvi, Prithvi-EO-2.0, SatMAE, and ScaleMAE pretrain large backbones on satellite imagery and then fine-tune the learned encoder to downstream dense EO prediction problems~\cite{jakubik2023foundation,roy2024prithvi,cong2022satmae,reed2023scalemae}. Many of these models follow a masked-autoencoding-style pretraining workflow: input images are divided into patches, projected into tokens by a \textit{patch-embedding layer}, processed by a \textit{vision transformer (ViT) encoder}, and reconstructed by a decoder used only during pretraining~\cite{dosovitskiy2021image,he2022masked}. After pretraining, the pretrained patch embedding and the encoder are reused for downstream fine-tuning, and a new decoder is added for downstream tasks such as segmentation. This pretrain-then-fine-tune workflow is efficient for EO because dense geospatial labels are expensive to obtain, and downstream datasets are often limited in geography, time, sensor type, or event coverage~\cite{mai2024opportunities,lacoste2023geobench,marsocci2026pangaea}. However, there are two challenges of adapting EO-pretrained GeoFMs to practical downstream datasets.

%One of the challenge EO-pretrained GeoFMs expose a fixed spectral interface. A pretrained patch embedding expects a particular set of input bands, while downstream datasets may come from different sensors, contain more bands, or include auxiliary channels. The common compatible baseline is \bandsel{}: choose the target bands whose wavelengths most closely match the pretrained input bands and discard the rest. This keeps the pretrained embedding intact, which is often important for stable transfer, but it can silently remove channels that carry task-relevant information. The opposite approach, replacing the embedding so that all target bands can be consumed directly, uses more information but introduces a new trainable spectral front end and can increase adaptation cost.

A first challenge is spectral mismatch. The pretrained patch embedding defines a fixed input band interface: it expects a particular number, order, and spectral meaning of input channels. However, downstream EO datasets may come from different sensors, or contain additional multispectral bands. A common compatibility strategy, denoted as \bandsel{} in this paper, is to select the downstream bands whose physical characteristics (such as wavelengths) best match the pretrained input bands, while discarding channels that cannot be matched. Similar band-matching strategies are used in recent GeoFM fine-tuning works to make heterogeneous downstream datasets compatible with fixed-interface pretrained models~\cite{marsocci2026pangaea}. This strategy preserves the pretrained patch embedding and provides a stable, physically motivated baseline. However, even when \bandsel{} selects a target band for a pretrained input slot, discarded channels may contain task-relevant information. An alternative is to redesign or adapt the input embedding so that heterogeneous or multisensor inputs can be consumed directly, as in wavelength-conditioned, multisensor, or any-sensor EO foundation models~\cite{xiong2024dofa,han2024msgfm,astruc2025anysat}. These approaches increase spectral flexibility, but they may also introduce additional trainable parameters during adaptation and thus increase adaptation cost.

\begin{figure}[t]
  \centering
  \includegraphics[width=\columnwidth]{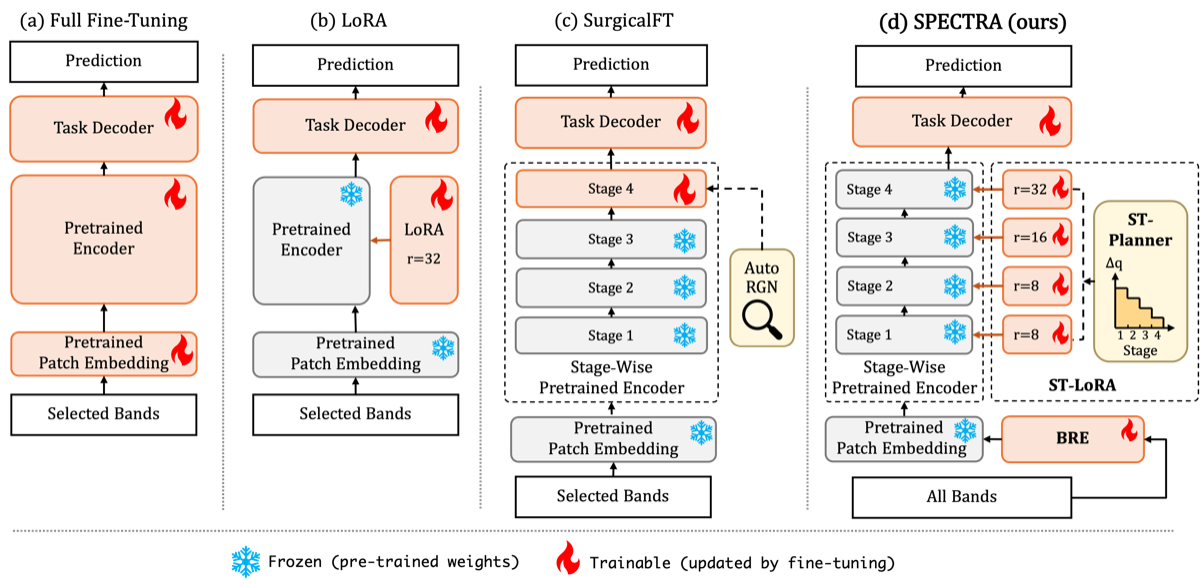}
  \Description{A schematic comparing fine-tuning policies, including frozen pretrained layers, trainable decoder components, BRE input adaptation, and ST-LoRA.}
  \caption{Comparison of PEFT methods compared in this work. \method{} adds a pretrained-compatible \bre{} input path and uses \starplanner{} to assign stage-wise LoRA ranks.}
  \label{fig:finetuning_comparison}
\end{figure}

A second challenge is fine-tuning cost. Different fine-tuning strategies can be different in computing cost. Full fine-tuning updates all parameters of a large pretrained encoder, which can be expensive in GPU memory, optimization time, and model storage. Parameter-efficient fine-tuning (PEFT) methods reduce this cost by freezing most pretrained weights and training only a small number of task-specific parameters~\cite{houlsby2019parameter}. LoRA, which injects low-rank trainable matrices into existing linear layers, ~\cite{hu2022lora}, is one of the most attractive PEFT methods because it is easy to implement and effective. Recent work on EO and GeoFM adaptation also shows that PEFT methods can provide competitive performance while reducing trainable parameters, training time, and memory requirements~\cite{dong2024upetu,marti2025fine}. However, standard LoRA can be further improved because it usually applies the same rank across layers or stages, even though different encoder stages may have different transferability to a target task. Another PEFT method, Surgical fine-tuning, shows that selectively adapting a subset of pretrained layers can improve transfer under distribution shift~\cite{lee2023surgical}, but unfreezing an entire stage can still require many more trainable parameters than LoRA. We compare full fine-tuning, LoRA, and Surgical fine-tuning and our proposed idea in Figure \ref{fig:finetuning_comparison}.

For both of the two challenges, although prior work has made remarkable progress on spectral adaptation and parameter-efficient fine-tuning separately, jointly addressing both challenges, to the best of our knowledge, still needs to be explored. A practical cross-sensor GeoFM PEFT method should use the spectral information available in downstream inputs. At the same time, it should adapt the encoder under a controlled trainable-parameter budget.

To handle the two problems, we propose \textbf{\method{}}, a parameter-efficient cross-sensor fine-tuning framework that solves the spectral mismatch challenge by input spectral adaptation and solves the cost challenge by stage-wise encoder adaptation. The framework contains two key components: Band-Routed Embedding (\bre{}) and Stage-wise Transferability-aware LoRA (\starplanner{}). \bre{} maps all available downstream bands into the band space expected by the pretrained model, producing a "virtual image" compatible to pre-training input. It preserves the selected-band input physically grounded anchor and learns routing-gated residual contributions from all fine-tuning bands, allowing information beyond the selected subset to refine the virtual pretrained-compatible image. Stage-wise Transferability-aware LoRA (\starplanner{}) addresses encoder adaptation cost by partitioning the pretrained encoder into stages and assigning stage-specific LoRA ranks before fine-tuning. It contains a transferability-aware planner (\textbf{ST planner}) that estimates stage-wise transferability using frozen-feature metrics such as LogME~\cite{you2021logme}, then allocates LoRA capacity based on the LogME profile. The resulting pipeline aims to learn which stage is useful to learn spectral information and assign trainable LoRA ranks accordingly, instead of using standard uniform LoRA ranks. 

% Our experiments evaluate \method{} across three EO-pretrained GeoFMs with different pretrained spectral configurations and four downstream segmentation datasets. The evaluation separates native or near-native settings, where \bandsel{} is already well aligned with the pretrained band interface, from mismatched settings where \bandsel{} creates an information bottleneck by discarding non-selected channels. Across five fine-tuning policies and 60 matched backbone--dataset--policy cells, adding \bre{} improves mean macro \miou{} by $+2.88$ points over the corresponding \bandsel{} baseline, whereas replacing \bandsel{} with a direct MLP projector decreases macro \miou{} by $-2.28$ points. The largest gains occur for ScaleMAE on FireScars, where the RGB-limited pretrained interface of ScaleMAE forces \bandsel{} to discard much of the multispectral downstream input. For example, on FireScars, \method{} reaches $82.05$ mIoU, compared with $68.20$ for LoRA and $72.30$ for full fine-tuning. On the other hand, for GeoFM--dataset pairs that are already aligned well by band-selection, such as Prithvi--FireScars, the improvement of \method{} is limited. For stage-wise encoder adaptation, completed Sen1Floods11 runs show that \method{} improves over the LoRA-32 \bandsel{} baseline for Prithvi, ScaleMAE, and SatMAE while reducing active trainable parameters by $17.1$--$33.6\%$. By comparing our proposed transferability-aware rank planner with multiple manually-selected rank schedules, we can find that \starplanner{} is an automatic low-budget rank allocation strategy.

Our experiments evaluate \method{} across three EO-pretrained GeoFMs and four downstream segmentation datasets, covering both spectrally aligned and mismatched transfer settings. Across five fine-tuning policies and 60 matched backbone--dataset--policy configurations, adding \bre{} improves mean macro \miou{} by $+2.88$ points over the corresponding \bandsel{} baseline, whereas a direct MLP projector underperforms \bandsel{}. The gains are most pronounced when band selection discards substantial downstream spectral information and are smaller when the pretrained and downstream band interfaces are already well aligned. In stage-wise adaptation on Sen1Floods11, \method{} also outperforms the LoRA-32 \bandsel{} baseline across all three backbones while reducing active trainable parameters by $17.1$--$33.6\%$. Comparisons with manually selected rank schedules further demonstrate that \starplanner{} provides effective automatic rank allocation under a limited parameter budget.

This paper makes three contributions with the proposed SPECTRA method:
\begin{itemize}
  % \item We propose \textbf{\textsc{SPECTRA}}, a cross-sensor parameter-efficient fine-tuning framework for GeoFMs. SPECTRA addresses two practical challenges: adapting different downstream spectral inputs to a fixed pretrained band interface, and reducing trainable-parameter budget compared to LoRA.

  \item We introduce \textbf{Band-Routed Embedding (\bre{})}, a lightweight input adaptation module that learns source-to-target band contributions for spectral mismatch between pre-training and fine-tuning datasets (downstream datasets). \bre{} learns routing-gated contributions from all fine-tuning bands, allowing information beyond the selected subset to refine the selected-band input without modifying the pretrained patch embedding.

  \item We introduce \textbf{Stage-wise Transferability-aware LoRA (\starplanner{})} for parameter-efficient GeoFM adaptation. \starplanner{} has an ST-planner that estimates the transferability of each pretrained encoder stage before fine-tuning and assigns stage-specific LoRA ranks. By concentrating trainable parameters in stages with high transferability and reducing parameters in other stages, \starplanner{} helps to reduce the number of trainable parameters compared to standard parameter-efficient LoRA.

  \item  We conduct a comprehensive evaluation across three EO-pretrained GeoFMs and four downstream segmentation datasets with different spectral configurations. The experiments include quantitative accuracy comparisons, trainable-parameter accounting, \bre{} ablation study, rank-schedule analyses, and qualitative visualization examples. Not only do the quantitative results show our methods are effective, but the visualization examples also illustrate that SPECTRA helps capture details in the images.

\end{itemize}

\section{Related Work}

\textit{Geospatial foundation models.}
Remote-sensing foundation models adapt masked autoencoding and transformer pretraining to imagery with spectral, spatial, and temporal structure. SatMAE introduces temporal and spectral positional structure for satellite imagery \cite{cong2022satmae}; ScaleMAE injects scale awareness for multiscale geospatial representation learning \cite{reed2023scalemae}; and Prithvi-style models pretrain large geospatial transformers on HLS-like imagery \cite{jakubik2023foundation,roy2024prithvi}. These models differ not only in architecture and pretraining data, but also in their assumed input sensors and native bands. \method{} targets this interface mismatch rather than treating the backbone as a generic image encoder.

\textit{Parameter-efficient fine-tuning.}
Fine-tuning all backbone parameters is a strong reference point, but it is expensive for large GeoFMs and can be difficult to justify when downstream labels are limited. LoRA injects low-rank updates into frozen layers and controls trainable capacity through the rank \cite{hu2022lora}. Uniform LoRA ranks are simple, but they assume every encoder stage needs the same adaptation capacity. \method{} keeps LoRA as the encoder adaptation mechanism and studies whether rank should instead vary by stage.
The differences of these approaches are illustrated in Figure~\ref{fig:finetuning_comparison}.% contrasts the policies: input adaptation, decoder-only tuning, uniform LoRA, and stage-wise LoRA.

\textit{Cross-sensor adaptation.}
Sensor mismatch is common in EO workflows: a downstream task may use a different satellite, a different subset of optical bands, SAR-derived channels, elevation features, or task-specific stacks. Band selection is robust because it keeps the pretrained interface unchanged, but it discards all other bands that might contain useful information. Learned projections and embedding replacement can ingest arbitrary bands; for example, a recent work~\cite{murphy2025deep} uses an MLP cross-sensor adapter for cloud-property retrieval. However, such adapters can also change the input distribution learned by the pretrained encoder. \bre{} is designed around a conservative principle: keep the pretrained patch embedding as the tokenizer and learn only a small virtual-band correction in image space.

\textit{Transferability diagnostics and rank allocation.}
Transferability metrics such as LEEP \cite{nguyen2020leep} and LogME \cite{you2021logme} estimate how useful pretrained features are before full fine-tuning. In our setting, these diagnostics serve two roles. First, they help explain why \bandsel{} performance differs across backbone-dataset pairs. Second, they provide a low-cost signal for allocating LoRA rank across encoder stages. \starplanner{} uses this diagnostic view for GeoFM fine-tuning, where both spectral mismatch and task transferability matter.

%%%%%%%%%%%%%%%%%%%%%%%%%%%%%%%%%%%%%%%%%%%%%%%%%%
%\section{Problem Statement}
\section{Background}
%\subsection{Problem Setup}
\subsection{Problem Statement}

Let $X \in \R^{C \times H \times W}$ denote a target image with $C$ available channels and dense labels $Y$. A pretrained GeoFM has a pre-trained patch embedding $E_0$ that expects $K$ compatible input bands. In cross-sensor fine-tuning, usually $C \ne K$ or the $C$ target channels do not match the spectral interface used during pretraining. The \bandsel{} baseline resolves this mismatch by selecting a subset $\mathcal{S}\subseteq\{1,\ldots,C\}$ with $|\mathcal{S}|=K$:
\begin{equation}
  X_{\mathrm{sel}} = P_{\mathrm{sel}}(X) \in \R^{K \times H \times W},
  \label{eq:bandsel}
\end{equation}
where $P_{\mathrm{sel}}$ selects the $K$ target channels whose central wavelengths or metadata best match the pretrained interface. This keeps the original patch embedding valid, but it makes every unselected channel invisible.

\textit{The first problem} is therefore spectral adaptation: construct a pretrained-compatible map that can use all observed target channels while still producing a $K$-channel input for $E_0$. We write this desired map as
\begin{equation}
  A_\theta: \R^{C \times H \times W}\rightarrow \R^{K \times H \times W},
  \qquad X_{\mathrm{all}} = A_\theta(X).
\end{equation}
In this paper, ``fully using the bands'' means that no target channel is hard-discarded by the input map. Formally, every channel can influence at least one virtual pretrained band. This condition does not claim that every band is always useful for every task; it states the structural requirement that all bands remain available to the fine-tuned model.

\textit{The second problem} is adaptation cost. Let $F_{\omega,\phi}$ denote a pretrained encoder with frozen backbone weights $\omega$ and trainable adaptation (e.g., LoRA weights) parameters $\phi$, and let $D_{\eta}$ denote the segmentation decoder with trainable parameters $\eta$. A fine-tuned prediction has the form $\hat{Y}=D\!\left(F_{\omega,\phi}\!\left(E_0(A_\theta(X))\right)\right)$. 
% \begin{equation}
%   \hat{Y}=D\!\left(F_{\omega,\phi}\!\left(E_0(A_\theta(X))\right)\right).
% \end{equation}
The training objective is standard supervised segmentation to minimize the loss $\mathcal{L}$ between prediction $\hat{Y}$ and true labels $Y$: $  \min_{\theta,\phi,\eta}\ \frac{1}{N}\sum_{n=1}^{N}
  \mathcal{L}\!\left(\hat{Y}_n,Y_n\right)$. 
% \begin{equation}
%   \min_{\theta,\phi,\eta}\ \frac{1}{N}\sum_{n=1}^{N}
%   \mathcal{L}\!\left(\hat{Y}_n,Y_n\right),
% \end{equation}
Our design goal is to keep the trainable percentage $\rho$ small. 
\begin{equation}
  \rho=\frac{|\theta|+|\phi|+|\eta|}
  {|\omega|+|\theta|+|\phi|+|\eta|},
  \label{eq:percent}
\end{equation}
To tackle the two problems, we \textbf{focus on improving $A_\theta$ and $\phi$}. \textbf{For the first problem}, we propose a new input adapter $A_\theta$. \textbf{For the second problem}, we propose a parameter-efficient fine-tuning method to decrease $\rho$ by reducing the number $\phi$. In this paper, \bandsel{} is the default input adapter $A_\theta$ for the baseline methods; \bre{} is our proposed all-band input adaptation module.

\subsection{Models and Datasets}

We use three GeoFMs and four downstream datasets to evaluate spectral transfer under different levels of input-band mismatch. Table~\ref{tab:geofm_backbones} summarizes the pretrained models. For each GeoFM, it reports the pretraining data source, the fixed input band interface, the number of bands $K$ expected by the pretrained patch embedding, and the ViT backbone size. Table~\ref{tab:datasets} summarizes the downstream datasets. For each dataset, it reports the input data source, the full number of available target channels $C$, the number of semantic classes, and the prediction task. Together, these tables define the 12 backbone--dataset combinations used in our experiments and show where spectral mismatch can arise. For example, Prithvi-EO-2.0 has a 6-band HLS optical interface, ScaleMAE has an RGB-only interface, and SatMAE has a 10-band Sentinel-2 multispectral interface, while the downstream datasets provide between 6 and 14 input channels. Full spectral-band metadata for the pretrained interfaces and downstream datasets are provided in Appendix Tables~\ref{tab:appendix_source_band_metadata} and~\ref{tab:appendix_target_band_metadata}.

\begin{table}[h]
  \centering
  \scriptsize
  \setlength{\tabcolsep}{2pt}
  \caption{GeoFM backbones, pretrained data sources, and spectral interfaces.}
  \label{tab:geofm_backbones}
  \begin{tabular}{@{}p{0.16\columnwidth}p{0.30\columnwidth}p{0.24\columnwidth}p{0.10\columnwidth}p{0.12\columnwidth}@{}}
    \toprule
    GeoFM & Pretrained data source & Input interface & Bands ($K$) & Backbone \\
    \midrule
    Prithvi-EO-2.0 & HLS (Landsat/Sentinel-2) \cite{roy2024prithvi} & HLS optical & 6 & ViT-Huge \\
    SatMAE & fMoW-Sentinel / Sentinel-2 \cite{cong2022satmae} & Sentinel-2 multispectral & 10 & ViT-Large \\
    ScaleMAE & fMoW/ImageNet RGB \cite{reed2023scalemae} & RGB & 3 & ViT-Large \\
    \bottomrule
  \end{tabular}
\end{table}

\begin{table}[h]
  \centering
  \scriptsize
  \setlength{\tabcolsep}{2pt}
  \caption{Downstream datasets, target input sources, and full target-channel counts.}
  \label{tab:datasets}
  \begin{tabular}{@{}p{0.18\columnwidth}p{0.34\columnwidth}p{0.10\columnwidth}p{0.09\columnwidth}p{0.18\columnwidth}@{}}
    \toprule
    Dataset & Target input source & Bands ($C$) & Classes & Task \\
    \midrule
    Sen1Floods11 & Sentinel-2 / Sen1Floods11 \cite{bonafilia2020sen1floods11} & 13 & 2 & Flood \\
    FireScars & HLS FireScars / GEO-Bench \cite{lacoste2023geobench} & 6 & 2 & Burn-scar \\
    Landslide4Sense & Sentinel-2 + DEM channels \cite{ghorbanzadeh2022landslide4sense} & 14 & 2 & Landslide \\
    SA Crop Type & Sentinel-2 L2A / GEO-Bench \cite{lacoste2023geobench} & 12 & 10 & Crop-type \\
    \bottomrule
  \end{tabular}
\end{table}

For all GeoFMs and datasets, the definition \bandsel{} method for our study is important because it is used in both baseline methods and our proposed adaptation module. We define the \bandsel{} method as Equation~\ref{eq:bandsel}. Also, Table~\ref{tab:spectral_transfer_diagnostics} summarizes how downstream target bands are mapped to the $K$ pretrained input slots for each GeoFM--dataset pair. The selected-index column lists the target channel indices fed to the pretrained patch embedding, in pretrained input-slot order. Repeated indices indicate that one target band is reused for multiple pretrained slots. Target bands that are not selected for any pretrained slot are marked as discarded for that GeoFM--dataset pair. Auxiliary channels, such as DEM, are not wavelength-matched and are treated as non-spectral auxiliary inputs. We also report a LogME-based \cite{you2021logme} transfer gap of the frozen pretrained encoder. Lower values indicate stronger transferability, suggesting that the frozen representation is already well aligned with the downstream task. Full band-name mappings are provided in Appendix Tables~\ref{tab:appendix_band_mapping_prithvi}, \ref{tab:appendix_band_mapping_scalemae}, and~\ref{tab:appendix_band_mapping_satmae}.

Together, these settings test three different spectral-transfer regimes using three GeoFMs. With \textbf{Prithvi-EO-2.0}, the downstream inputs are close to the HLS optical interface used during pretraining. In this regime, \bandsel{} can select $K$ highly matched channels from the target input $X$, and \bre{} mainly tests whether the remaining fine-tuning bands provide useful residual information beyond this strong selected-band baseline. Prithvi--FireScars is the clearest native-match case, since both the pretrained interface and the downstream dataset use six HLS-like optical bands. With \textbf{ScaleMAE}, the pretrained interface is much narrower because the pre-trained model expects only RGB inputs, and the RGB wavelengths are not always exactly aligned with the target multispectral bands. This creates an information bottleneck for \bandsel{} because it must discard most non-RGB channels; \bre{} is therefore expected to allow all fine-tuning bands to contribute to the input. With \textbf{SatMAE}, the pretrained interface is richer than RGB but still not perfectly matched to every downstream sensor interface. Although \bandsel{} maps the target channels to the pretrained $K$ slots using the closest available central wavelengths, some slots may only match approximately and even have repeated target bands, leaving a larger spectral gap for \bre{} to compensate. These three cases therefore evaluate \bre{} under native HLS-compatible, RGB-limited, and multispectral-but-misaligned transfer settings.

\begin{table}[t]
  \centering
  \scriptsize
  \setlength{\tabcolsep}{1pt}
  \renewcommand{\arraystretch}{1.00}
  \caption{Band-selection and transferability diagnostics for all GeoFM--dataset pairs in this study.}
  \label{tab:spectral_transfer_diagnostics}
  \begin{tabular}{@{}>{\raggedright\arraybackslash}p{0.11\columnwidth}
                  >{\raggedright\arraybackslash}p{0.18\columnwidth}
                  >{\raggedright\arraybackslash}p{0.28\columnwidth}
                  >{\raggedright\arraybackslash}p{0.31\columnwidth}
                  p{0.09\columnwidth}@{}}
    \toprule
    Backbone & Dataset & Selected band indices & Discarded band indexes
 & Transfer gap $\downarrow$ \\
    \midrule

    \multirow{4}{*}{Prithvi}
      & {Sen1Floods11}
      & {[1, 2, 3, 8, 11, 12]}
      & {[0, 4, 5, 6, 7, 9, 10]}
      & {0.223} \\
      & {FireScars}
      & {[0, 1, 2, 3, 4, 5]}
      & {--}
      & {0.401} \\
      & {Landslide4Sense}
      & {[1, 2, 3, 8, 11, 12]}
      & {[0, 4, 5, 6, 7, 9, 10, 13]}
      & {0.338} \\
      & {SA Crop Type}
      & {[1, 2, 3, 8, 10, 11]}
      & {[0, 4, 5, 6, 7, 9]}
      & {0.545} \\
    \midrule

    \multirow{4}{*}{ScaleMAE}
      & {Sen1Floods11}
      & {[3, 2, 1]}
      & {[0, 4, 5, 6, 7, 8, 9, 10, 11, 12]}
      & {0.555} \\
      & {FireScars}
      & {[2, 1, 0]}
      & {[3, 4, 5]}
      & {0.748} \\
      & {Landslide4Sense}
      & {[3, 2, 1]}
      & {[0, 4, 5, 6, 7, 8, 9, 10, 11, 12, 13]}
      & {0.000} \\
      & {SA Crop Type}
      & {[3, 2, 1]}
      & {[0, 4, 5, 6, 7, 8, 9, 10, 11]}
      & {0.541} \\
    \midrule

    \multirow{4}{*}{SatMAE}
      & {Sen1Floods11}
      & {[1, 2, 3, 4, 5, 6, 7, 8, 11, 12]}
      & {[0, 9, 10]}
      & {0.194} \\
      & {FireScars}
      & {[0, 1, 2, 2, 2, 3, 3, 3, 4, 5]}
      & {--}
      & {0.531} \\
      & {Landslide4Sense}
      & {[1, 2, 3, 4, 5, 6, 7, 8, 11, 12]}
      & {[0, 9, 10, 13]}
      & {0.203} \\
      & {SA Crop Type}
      & {[1, 2, 3, 4, 5, 6, 7, 8, 10, 11]}
      & {[0, 9]}
      & {0.580} \\
    \bottomrule
  \end{tabular}
\end{table}

% Together, these settings test three different spectral-transfer regimes using three GeoFMs. With Prithvi-EO-2.0, the downstream inputs are close to the HLS optical interface used during pretraining. In this regime, \bandsel{} can select $K$ highly matched channels from the target input $X$, and \bre{} mainly tests whether the remaining fine-tuning bands provide useful residual information beyond this strong selected-band baseline. Prithvi--FireScars is the clearest native-match case, since both the pretrained interface and the downstream dataset use six HLS-like optical bands. With ScaleMAE, the pretrained interface is much narrower because the pre-trained model expects only RGB inputs, and the RGB wavelengths are not always exactly aligned with the target multispectral bands. This creates an information bottleneck for \bandsel{} because it must discard most non-RGB channels; \bre{} is therefore expected to allows all fine-tuning bands to contribute to the input. With SatMAE, the pretrained interface is richer than RGB but still not perfectly matched to every downstream sensor interface. Although \bandsel{} maps the target channels to the pretrained $K$ slots using the closest available central wavelengths, some slots may only match approximately and even have repeated target bands, leaving a larger spectral gap for \bre{} to compensate. These three cases therefore evaluate \bre{} under native HLS-compatible, RGB-limited, and multispectral-but-misaligned transfer settings.

\section{Method}

\begin{figure*}[t]
  \centering
  \includegraphics[width=\textwidth]{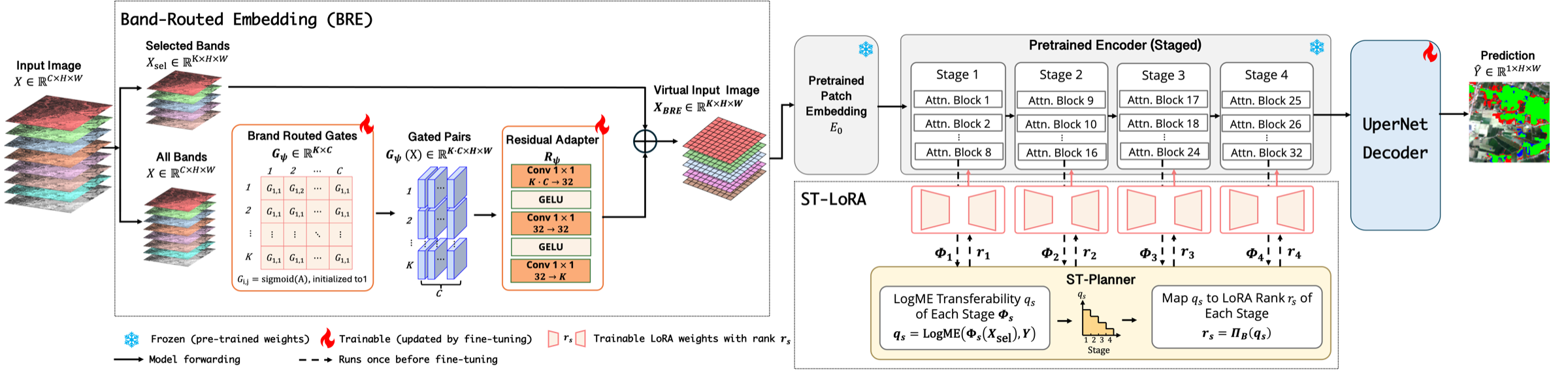}
  % \Description{A model architecture diagram for SPECTRA. All target bands enter band-routed gates and a residual adapter to form virtual pretrained-compatible bands. The frozen pretrained patch embedding feeds a staged encoder with LoRA modules. A ST-planner computes stage transferability scores and maps them to LoRA ranks. A UPerNet decoder outputs the segmentation map.}
  % \caption{\method{} model architecture. \bre{} routes all target bands into a pretrained-compatible virtual input before the frozen patch embedding, while \starplanner{} assigns stage-wise LoRA ranks from transferability scores. The architecture is designed to use bands discarded by \bandsel{} while controlling trainable adaptation cost.}
    \Description{A model architecture diagram for SPECTRA. All target bands enter band-routed gates and a residual adapter to form virtual pretrained-compatible bands. The frozen pretrained patch embedding feeds a staged encoder with LoRA modules. A stage-wise transferability planner computes transferability scores and maps them to LoRA ranks. A UPerNet decoder outputs the segmentation map.}
  % \caption{\method{} architecture. \bre{} converts the full target image into a $K$-band pretrained-compatible virtual image before the frozen patch embedding, allowing bands discarded by \bandsel{} to contribute through a learned residual path. \starplanner{} runs once before fine-tuning and assigns stage-wise LoRA ranks based on transferability metrics from a frozen encoder. The resulting pipeline addresses spectral mismatch while controlling trainable adaptation cost.}
  \caption{\method{} architecture. \bre{} maps a target image with $C$ observed bands to a $K$-band pretrained-compatible virtual image by combining a selected-band path with a learned all-band residual path. \starplanner{} runs once before fine-tuning to estimate transferability using frozen encoder features and assign stage-wise LoRA ranks based. Together, they handle spectral mismatch while controlling adaptation cost.}
  % \caption{\method{} architecture. \bre{} maps the full target image to a $K$-band pretrained-compatible virtual image before the frozen patch embedding, while \starplanner{} assigns stage-wise LoRA ranks based on transferability scores before fine-tuning starts. }

  \label{fig:method_overview}
\end{figure*}

\method{} has two proposed components: \bre{} and \starplanner{}, and Figure~\ref{fig:method_overview} illustrates how the two components work with other modules in the fine-tuning pipeline. To begin with, \bre{} converts the full target image $X\in\R^{C\times H\times W}$ into a virtual input image $X_{BRE}\in\R^{K\times H\times W}$ that is compatible with $K$-band pretrained input images. It starts from the \bandsel{} anchor, learns band-routed gates over all observed bands, and applies a residual adapter to learn a projection so that the resulting $X_{BRE}$ is the same shape as the input image expected by the pretrained patch embedding $E_0$ of the GeoFM. The virtual input image is learned by $E_0$ embedded tokens for the pretrained encoder of the GeoFM. The pretrained encoder is divided into several (e.g. four) stages, and each encoder stage is adapted by LoRA which is also divided into the same stages, where each stage has its own LoRA rank. For the stage-wise LoRA, \starplanner{} estimates stage-wise transferability to estimate stage-wise LoRA ranks for the encoder. After getting the embedded features from the encoder, the prediction is finally produced by the UPerNet decoder. 

\subsection{Band-Routed Embedding}

\bre{} is our input adaptation module for spectral mismatch. As defined in the problem statement, the goal of input adaptation is to construct a map
$A_\theta:\R^{C\times H\times W}\rightarrow\R^{K\times H\times W}$
that uses the full target image $X$ while still producing a $K$-channel image compatible with the pretrained patch embedding $E_0$. In our method, \bre{} implements this adapter. We denote its output as 
\[
  X_{\mathrm{BRE}} = A_\theta(X) \in \R^{K\times H\times W}.
\]
As illustrated in Figure~\ref{fig:method_overview}, there are two paths derived from the input images: a selected-band path (the upper path in the figure), and a residual path (the lower path in the figure) adapting all bands as input with band routed gates and a residual adapter. The key design principle is to keep the selected-band input as a physically grounded anchor and to add a learned residual correction from all available target bands.

\paragraph{Selected-band anchor.}
Let $\mathcal{S}=(s_1,\ldots,s_K)$ be the ordered band indices used by \bandsel{}, and $X_{\mathrm{S}}$ is the input image required by the pretrained patch embedding layer of the GeoFM. The selected-band image is $  X_{\mathrm{sel}} = P_{\mathrm{sel}}(X)$, where the order follows the pretrained input-band slots: $X_{\mathrm{sel},j}=X_{s_j}$. This anchor preserves the standard \bandsel{} path and keeps the input close to the distribution expected by the pretrained patch embedding.

\paragraph{All-band residual path.}
From this path, \bre{} learns a residual correction $\Delta_\theta(X)$ in the same $K$-band space to be added to the selected $K$ bands:
\begin{equation}
  X_{\mathrm{BRE}} = X_{\mathrm{sel}} + \Delta_\theta(X),
  \label{eq:bre_residual}
\end{equation}
where $\Delta_\theta(X)\in\R^{K\times H\times W}$. The residual path is zero-initialized at its final layer, so $\Delta_\theta(X)=0$ at model initialization. Therefore, the initial forward pass of \bre{} is exactly the same as \bandsel{}: $X_{\mathrm{BRE}} = X_{\mathrm{sel}}$. This makes fine-tuning start from a stable selected-band baseline instead of an unconstrained learned projection.

\paragraph{Band routed gates.}
To allow every band in the fine-tuning dataset (shortened as "\textbf{fine-tuning band}" below) to contribute to every virtual pretrained band, the all-band residual path uses a learned gate table $G_\psi \in \R^{K\times C}$.

Each row $G_{j,:}$ corresponds to one virtual pretrained band $j$, and each entry $G_{j,i}$ controls the contribution of inpu band $i$ to that band fo the virtual image (shortened as "\textbf{virtual band}" below). The gates are independent sigmoid gates:
\begin{equation}
  G_{j,i}=2\,\mathrm{sigmoid}(L_{j,i}),
  \qquad L\in\R^{K\times C}.
\end{equation}
where $L_{j,i}$ is the learnable logit for the gate connecting the fine-tuning band $i$ to virtual band $j$. We initialize $L=0$, so every gate starts from $G_{j,i}=1$. Thus, the gates initially preserve all source-band signals rather than suppressing them.

The output of band routed gates is gated paris (as illustrated in Figure~\ref{fig:method_overview}). Gated pairs save the contribution of observed source band $c$ to the residual correction of virtual pretrained band $k$. For each virtual band $j$, the gated fine-tuning bands are formed by multiplying all $C$ input channels with the corresponding gate row:
\begin{equation}
  U_j = \mathrm{concat}_{i=1}^{C}\left(G_{j,i}X_i\right)
  \in \R^{C\times H\times W}.
\end{equation}
The gated pairs for all $K$ virtual bands are then concatenated:
\[
  U = \mathrm{concat}_{j=1}^{K} U_j
  \in \R^{KC\times H\times W}.
\]
This construction ensures that no target channel is hard-discarded: every observed fine-tuning band can influence every virtual pretrained band through the residual path.

\paragraph{Shared residual adapter.}
To finally output the virtual image correction $\Delta_\theta(X)$ with $K$ bands, the concatenated gated tensor $U$ is passed to one shared lightweight residual adapter $  \Delta_\theta(X)=R_\theta(U)$, where $R_\theta$ is a compact $1\times1$ convolutional network as shown in Figure \ref{fig:method_overview}. This is a single shared adapter over the concatenated $KC$ gated channels, not $K$ separate target-wise adapters.
% \[
%   \Delta_\theta(X)
%   =
%   R_\theta(U),
% \]

% \[
%   \mathrm{Conv}_{1\times1}(KC\!\rightarrow\!32)
%   \rightarrow
%   \mathrm{GELU}
%   \rightarrow
%   \mathrm{Conv}_{1\times1}(32\!\rightarrow\!32)
%   \rightarrow
%   \mathrm{GELU}
%   \rightarrow
%   \mathrm{Conv}_{1\times1}(32\!\rightarrow\!K).
% \]
% \[
% \begin{aligned}
% R_{\theta} ={}&
% \mathrm{Conv}_{1\times1}(KC\!\rightarrow\!32)
% \rightarrow \mathrm{GELU}
% \rightarrow \mathrm{Conv}_{1\times1}(32\!\rightarrow\!32) \\
% &\rightarrow \mathrm{GELU}
% \rightarrow \mathrm{Conv}_{1\times1}(32\!\rightarrow\!K).
% \end{aligned}
% \]

\paragraph{Connection to the pretrained GeoFM.} After \bre{} forms the virtual image $X_{\mathrm{BRE}}$, the original pretrained patch embedding tokenizes it $Z = E_0(X_{\mathrm{BRE}})$, where $Z\in\R^{N_p\times d}$ is the patch-token sequence with ViT encoder depth $d$ and the number of patches after embedding \cite{dosovitskiy2020image}. Since $X_{\mathrm{BRE}}$ has exactly $K$ channels, the pretrained patch embedding receives the same number of input channels as during pretraining. Thus, \bre{} uses all available downstream bands without modifying the pretrained patch-embedding layers.

\subsection{Stage-wise Transfer-aware LoRA Rank Planning}
\label{sec:star}

 Our proposed \starplanner{}, a stage-wise transfer-aware rank planner that runs once before fine-tuning and assigns a LoRA rank to each encoder stage. The resulting rank schedule is then fixed during training.
Importantly, \starplanner{} does not route spectral bands or process image
features during fine-tuning.

Let the frozen pretrained encoder be divided into $S$ stages,
\begin{equation}
  F_{\omega}
  =
  F_{\omega}^{(S)} \circ \cdots \circ F_{\omega}^{(1)},
\end{equation}
where $\omega$ denotes the pretrained encoder weights. To profile stage-wise
transferability, \starplanner{} uses the selected-band path from
\bre{} without applying the gated redisual path. Specifically, the selected bands are embedded by
the original pretrained patch embedding:
\begin{equation}
  Z_{\mathrm{sel}} = E_0(X_{\mathrm{sel}}), \qquad
  H_0 = Z_{\mathrm{sel}}, \qquad
  H_s = F_{\omega}^{(s)}(H_{s-1}), \quad s=1,\ldots,S .
\end{equation}

For each stage $s$, we construct a feature matrix
$\Phi_s(X_{\mathrm{sel}})$ from the patch tokens in $H_s$, and then $H_s$ goes to the next stage $H_{s+1}$. The class label token is removed. Patch tokens are aligned with the segmentation label map by
the patch grid, and only patches whose dominant-class purity exceeds a threshold
are retained. The retained patch features are flattened over the profiling
images. We then measure stage-wise transferability using LogME~\cite{you2021logme}, which estimates the label evidence given frozen features without fine-tuning. The transferability score of stage $s$ is
\begin{equation}
  q_s =
  \mathrm{LogME}\!\left(\Phi_s(X_{\mathrm{sel}}), Y\right),
\end{equation}
where $Y$ denotes the profiling labels. The vector
$q=(q_1,\ldots,q_S)$ summarizes how useful the frozen representation at each stage is for the target task. Higher values means the corresponding stage learns better for the specific downstream task than other stages.

\paragraph{Budget-controlled rank formulation.}
Now we have the transferability of each stage $q_s$, and the next step is to calculate the LoRA rank for each stage with limit computing cost. To control the computing cost of ST-LoRA, there needs to be an upper bound for LoRA rank assigned to each stage. For this purpose, we define a reference uniform rank $r_{\mathrm{ref}}$. A uniform LoRA-$r_{\mathrm{ref}}$ baseline assigns $r^{\mathrm{uni}}
  =
  (r_{\mathrm{ref}},\ldots,r_{\mathrm{ref}})$
to all $S$ stages, so this is the standard LoRA method. We define the corresponding \textbf{total stage-rank budget} as
\begin{equation}
  T_{\mathrm{rank}}(r_{\mathrm{ref}})
  =
  \sum_{s=1}^{S} r_{\mathrm{ref}}
  =
  S r_{\mathrm{ref}} .
\end{equation}
This quantity is used as an architecture-level rank budget throughout the paper.
For example, with $S=4$ stages and $r_{\mathrm{ref}}=32$, the total rank budget matched to
uniform LoRA-32 is $T_{\mathrm{rank}}(32)=4\times 32=128$.

Thus, \starplanner{} is allowed to redistribute the same \textbf{total rank budget across}
stages:
\begin{equation}
  r_s \in \mathcal{R}(r_{\mathrm{ref}}), \qquad
  \sum_{s=1}^{S} r_s \le T_{\mathrm{rank}}(r_{\mathrm{ref}}).
\end{equation}
The range of \textbf{stage-wise rank} is defined as
\begin{equation}
  \mathcal{R}(r_{\mathrm{ref}})
  =
  \{0,4,8,\ldots,r_{\mathrm{ref}}/2,r_{\mathrm{ref}},2r_{\mathrm{ref}}\}.
\end{equation}
For instance, $\mathcal{R}(32)=\{0,4,8,16,32,64\}$.
Please note that $T_{\mathrm{rank}}$ is not a parameter count; exact trainable parameter
counts are computed and reported in the experiments.

\paragraph{Transfer and repair planning modes.}
Given the transferability profile $q$, \starplanner{} supports two complementary
planning modes. The \emph{transfer} mode assumes that stages that are already linearly useful for the target
task if it has high transferability. So it allocates more rank to stages with
higher transferability scores:
\begin{equation}
  a_s = q_s .
\end{equation}

The \emph{repair} mode assumes that stages with lower transferability need more corrective
LoRA capacity. So it allocates more rank to stages with larger transferability
gaps. The repair signal is 
\begin{equation}
  a_s = \max_{s} q_s - q_s .
\end{equation}
% where $q^{\star} = \max_{s} q_s$
% \begin{equation}
%   a_s = d_s .
% \end{equation}

For either mode, the stage signal $a_s$ is combined and normalized with temperature $\tau$:
\begin{equation}
  w_s
  =
  \frac{ \exp(a_s/\tau)}
       {\sum_{t=1}^{S} \exp(a_t/\tau)} .
\end{equation}
The temperature $\tau$ controls how concentrated the allocation is across stages. 

\paragraph{Discrete rank allocation.} This step estimates stage-wise LoRA rank for both of the transfer and repair modes using score $a_s $ with controlled rank budget. Using the budget-matched total rank budget, \starplanner{} first assigns
continuous stage ranks
\begin{equation}
  \tilde r_s
  =
  T_{\mathrm{rank}}(r_{\mathrm{ref}}) w_s .
\end{equation}
Each $\tilde r_s$ is quantized to the largest value in
$\mathcal{R}(r_{\mathrm{ref}})$ that does not exceed it:
\begin{equation}
  r_s^{(0)}
  =
  \max
  \left\{
  r \in \mathcal{R}(r_{\mathrm{ref}})
  :
  r \le \tilde r_s
  \right\}.
\end{equation}
The remaining budget is then used to greedily upgrade stages when the upgrade
reduces the quantization error without violating the total budget. The final
stage-wise rank schedule is therefore
\begin{equation}
  r=\Pi_{T_{\mathrm{rank}}}
  \left(q; r_{\mathrm{ref}}, \mathrm{mode}\right),
  \hspace{0.5em}
  r_s \in \mathcal{R}(r_{\mathrm{ref}}),
  \hspace{0.5em}
  \sum_{s=1}^{S} r_s \le T_{\mathrm{rank}}(r_{\mathrm{ref}}).
\end{equation}
Because the constraint is an upper bound, the discretized schedule can use fewer
active rank units than the uniform LoRA-$r_{\mathrm{ref}}$ baseline, which can
reduce trainable parameters and computation while remaining budget-matched to
the reference baseline.

\paragraph{Nested LoRA with stage-wise ranks.}
After planning, every adapted linear layer $\ell$ in stage $s$ uses the selected
rank $r_s$. Let $W_{\ell}$ and $b_{\ell}$ be the frozen pretrained weight and
bias of layer $\ell$. We use nested LoRA so that a maximum-rank adapter can be
truncated to different ranks at different stages:
\begin{equation}
  \ell(x)
  =
  W_{\ell}x + b_{\ell}
  +
  \mathbbm{1}[r_s>0]\,
  \frac{\alpha}{r_{\max}}
  B_{\ell}^{(r_s)} A_{\ell}^{(r_s)} x .
\end{equation}
Here $r_{\max}=2r_{\mathrm{ref}}$ is the largest rank in the range $\mathcal{R}(r_{\mathrm{ref}})$,
$A_{\ell}^{(r_s)}$ keeps the first $r_s$ rows of the nested LoRA matrix
$A_{\ell}$, and $B_{\ell}^{(r_s)}$ keeps the first $r_s$ columns of
$B_{\ell}$. Both matrix $A_{\ell}$, and $B_{\ell}^{(r_s)}$ are trainable LoRA weights defined by original LoRA paper \cite{hu2022lora}. If $r_s=0$, the LoRA residual for that stage is disabled, and only the frozen encoder parameter exist for that stage. The
pretrained encoder weights $W_{\ell}$ and $b_{\ell}$ remain frozen; only the
active LoRA parameters are trained. The \bre{} module and the segmentation
decoder are trained normally.

In our experiments, we evaluate both \emph{transfer} and \emph{repair} planning
modes under the same reference rank $r_{\mathrm{ref}}$. Unless otherwise
specified, the reported \method{} result uses the better of the two modes
selected by validation performance.

\begin{table*}[p]
  \centering
  \fontsize{6.3}{5.05}\selectfont
  \setlength{\tabcolsep}{1.4pt}
  \renewcommand{\arraystretch}{0.88}
  \caption{Full main comparison across backbones and datasets. Full-FT and linear probing (LP) serve as high- and low-cost references, respectively. Uniform LoRA, Surgical, and \method{} are different PEFT methods. Bold indicates the best mIoU for each GeoFM--dataset pair, excluding Full-FT.}
  \label{tab:main_comparison}
  \begin{tabular}{@{}p{0.12\textwidth}p{0.10\textwidth}p{0.12\textwidth}p{0.15\textwidth}p{0.17\textwidth}p{0.12\textwidth}@{}}
\toprule
Dataset & Method & Trainable params. \%$\downarrow$ & Test macro \miou{} (\%)$\uparrow$  & Test macro F1 (\%)$\uparrow$  & FG IoU (\%)$\uparrow$  \\
\midrule
\multicolumn{6}{@{}l}{\textbf{GeoFM: Prithvi-EO-2.0 600M}} \\
\midrule
\multirow{7}{=}{Sen1Floods11} & LP & 2.88\% & $85.91 \pm 0.75$ & $92.04 \pm 0.49$ & $74.95 \pm 1.55$ \\
 & LoRA-16 & 4.50\% & $84.46 \pm 1.91$ & $91.09 \pm 1.24$ & $72.38 \pm 3.22$ \\
 & LoRA-32 & 6.12\% & $85.10 \pm 1.70$ & $91.51 \pm 1.09$ & $73.52 \pm 2.75$ \\
 & LoRA-64 & 9.35\% & $84.40 \pm 3.18$ & $91.05 \pm 2.07$ & $72.50 \pm 5.13$ \\
 & Surgical & 27.16\% & $84.81 \pm 0.27$ & $91.34 \pm 0.19$ & $73.06 \pm 0.64$ \\
 & Full-FT & 100.00\% & $86.24 \pm 2.95$ & $92.23 \pm 1.83$ & $75.56 \pm 5.09$ \\
 & \textbf{SPECTRA} & 4.06\% & $\mathbf{86.39 \pm 0.85}$ & $\mathbf{92.34 \pm 0.53}$ & $\mathbf{75.74 \pm 1.35}$ \\
\midrule
\multirow{7}{=}{FireScars} & LP & 2.88\% & $84.27 \pm 0.45$ & $91.41 \pm 0.27$ & $80.30 \pm 0.66$ \\
 & LoRA-16 & 4.50\% & $85.25 \pm 0.35$ & $91.99 \pm 0.21$ & $81.67 \pm 0.45$ \\
 & LoRA-32 & 6.12\% & $85.47 \pm 0.35$ & $92.13 \pm 0.20$ & $81.98 \pm 0.41$ \\
 & \textbf{LoRA-64} & 9.35\% & $\mathbf{85.56 \pm 0.45}$ & $\mathbf{92.18 \pm 0.26}$ & $\mathbf{82.10 \pm 0.53}$ \\
 & Surgical & 27.16\% & $85.52 \pm 0.58$ & $92.16 \pm 0.34$ & $82.01 \pm 0.71$ \\
 & Full-FT & 100.00\% & $87.71 \pm 1.13$ & $93.42 \pm 0.65$ & $84.65 \pm 1.32$ \\
 & SPECTRA & 5.31\% & $85.40 \pm 0.33$ & $92.08 \pm 0.19$ & $81.75 \pm 0.33$ \\
\midrule
\multirow{7}{=}{Landslide4Sense} & LP & 2.88\% & $75.33 \pm 0.11$ & $83.96 \pm 0.09$ & $43.15 \pm 0.42$ \\
 & LoRA-16 & 4.50\% & $76.88 \pm 0.39$ & $85.22 \pm 0.32$ & $55.31 \pm 0.75$ \\
 & LoRA-32 & 6.12\% & $77.07 \pm 0.05$ & $85.38 \pm 0.04$ & $53.06 \pm 0.06$ \\
 & LoRA-64 & 9.35\% & $77.00 \pm 0.30$ & $85.33 \pm 0.23$ & $55.62 \pm 0.53$ \\
 & Surgical & 27.16\% & $76.32 \pm 0.17$ & $84.76 \pm 0.14$ & $52.40 \pm 0.87$ \\
 & Full-FT & 100.00\% & $77.25 \pm 0.08$ & $85.52 \pm 0.07$ & $54.49 \pm 0.40$ \\
 & \textbf{SPECTRA} & 3.29\% & $\mathbf{77.28 \pm 0.09}$ & $\mathbf{85.55 \pm 0.08}$ & $\mathbf{56.11 \pm 0.18}$ \\
\midrule
\multirow{7}{=}{SA Crop Type} & LP & 2.88\% & $32.86 \pm 0.13$ & $46.32 \pm 0.13$ & $31.51 \pm 0.63$ \\
 & LoRA-16 & 4.50\% & $35.46 \pm 0.53$ & $49.08 \pm 0.59$ & $33.27 \pm 1.59$ \\
 & LoRA-32 & 6.12\% & $35.61 \pm 0.56$ & $49.22 \pm 0.70$ & $34.04 \pm 0.44$ \\
 & LoRA-64 & 9.35\% & $35.87 \pm 0.14$ & $49.53 \pm 0.26$ & $33.42 \pm 0.12$ \\
 & Surgical & 27.16\% & $34.08 \pm 0.44$ & $47.73 \pm 0.51$ & $33.08 \pm 0.88$ \\
 & Full-FT & 100.00\% & $37.63 \pm 1.01$ & $51.65 \pm 1.10$ & $36.96 \pm 1.58$ \\
 & S\textbf{PECTRA} & 5.31\% & $\mathbf{36.32 \pm 0.63}$ & $\mathbf{49.96 \pm 0.75}$ & $\mathbf{33.58 \pm 0.61}$ \\
\midrule
\multicolumn{6}{@{}l}{\textbf{GeoFM: ScaleMAE}} \\
\midrule
\multirow{7}{=}{Sen1Floods11} & LP & 4.75\% & $70.16 \pm 4.12$ & $80.35 \pm 3.50$ & $48.29 \pm 6.80$ \\
 & LoRA-16 & 6.73\% & $76.27 \pm 2.21$ & $85.25 \pm 1.71$ & $57.92 \pm 4.18$ \\
 & LoRA-32 & 8.72\% & $77.07 \pm 2.05$ & $85.85 \pm 1.58$ & $59.17 \pm 3.80$ \\
 & LoRA-64 & 12.68\% & $77.17 \pm 1.76$ & $85.93 \pm 1.33$ & $59.34 \pm 3.18$ \\
 & Surgical & 28.57\% & $73.39 \pm 2.46$ & $82.98 \pm 1.99$ & $52.98 \pm 4.39$ \\
 & Full-FT & 100.00\% & $75.41 \pm 0.99$ & $84.64 \pm 0.80$ & $56.54 \pm 1.93$ \\
 & \textbf{SPECTRA} & 7.23\% & $\mathbf{85.45 \pm 2.08}$ & $\mathbf{91.74 \pm 1.32}$ & $\mathbf{74.11 \pm 3.50}$ \\
\midrule
\multirow{7}{=}{FireScars} & LP & 4.75\% & $67.78 \pm 1.12$ & $80.49 \pm 0.77$ & $59.31 \pm 0.85$ \\
 & LoRA-16 & 6.73\% & $67.86 \pm 1.96$ & $80.62 \pm 1.38$ & $60.64 \pm 1.60$ \\
 & LoRA-32 & 8.72\% & $68.20 \pm 1.89$ & $80.86 \pm 1.31$ & $61.04 \pm 1.29$ \\
 & LoRA-64 & 12.68\% & $69.03 \pm 0.91$ & $81.44 \pm 0.61$ & $61.58 \pm 0.40$ \\
 & Surgical & 28.57\% & $68.27 \pm 1.56$ & $80.90 \pm 1.09$ & $60.72 \pm 1.31$ \\
 & Full-FT & 100.00\% & $72.30 \pm 1.39$ & $83.72 \pm 0.92$ & $65.24 \pm 1.17$ \\
 & \textbf{SPECTRA} & 7.48\% & $\mathbf{82.05 \pm 2.07}$ & $\mathbf{90.05 \pm 1.29}$ & $\mathbf{77.36 \pm 2.94}$ \\
\midrule
\multirow{7}{=}{Landslide4Sense} & LP & 4.75\% & $74.13 \pm 0.05$ & $82.93 \pm 0.04$ & $50.18 \pm 0.11$ \\
 & LoRA-16 & 6.73\% & $75.05 \pm 0.20$ & $83.71 \pm 0.17$ & $51.89 \pm 0.37$ \\
 & LoRA-32 & 8.72\% & $75.16 \pm 0.28$ & $83.81 \pm 0.23$ & $52.12 \pm 0.49$ \\
 & LoRA-64 & 12.68\% & $75.15 \pm 0.19$ & $83.79 \pm 0.16$ & $52.04 \pm 0.33$ \\
 & Surgical & 28.57\% & $74.82 \pm 0.31$ & $83.50 \pm 0.26$ & $51.40 \pm 0.56$ \\
 & Full-FT & 100.00\% & $75.36 \pm 0.07$ & $83.97 \pm 0.06$ & $52.44 \pm 0.14$ \\
 & \textbf{SPECTRA} & 6.20\% & $\mathbf{76.33 \pm 0.06}$ & $\mathbf{84.77 \pm 0.06}$ & $\mathbf{54.25 \pm 0.15}$ \\
\midrule
\multirow{7}{=}{SA Crop Type} & LP & 4.75\% & $30.36 \pm 0.42$ & $43.37 \pm 0.71$ & $24.44 \pm 2.42$ \\
 & LoRA-16 & 6.73\% & $31.50 \pm 0.44$ & $44.39 \pm 0.60$ & $26.35 \pm 1.22$ \\
 & LoRA-32 & 8.72\% & $31.90 \pm 0.34$ & $45.08 \pm 0.58$ & $25.91 \pm 2.16$ \\
 & LoRA-64 & 12.68\% & $32.09 \pm 0.35$ & $45.06 \pm 0.46$ & $26.74 \pm 1.25$ \\
 & Surgical & 28.57\% & $31.03 \pm 0.33$ & $44.02 \pm 0.27$ & $25.44 \pm 1.58$ \\
 & Full-FT & 100.00\% & $33.22 \pm 0.52$ & $46.58 \pm 0.63$ & $29.26 \pm 1.84$ \\
 & \textbf{SPECTRA} & 7.56\% & $\mathbf{31.97 \pm 0.30}$ & $\mathbf{45.03 \pm 0.36}$ & $\mathbf{29.57 \pm 1.55}$ \\
\midrule
\multicolumn{6}{@{}l}{\textbf{GeoFM: SatMAE}} \\
\midrule
\multirow{7}{=}{Sen1Floods11} & LP & 4.75\% & $80.83 \pm 5.01$ & $88.57 \pm 3.43$ & $66.21 \pm 8.52$ \\
 & LoRA-16 & 6.73\% & $83.76 \pm 1.45$ & $90.63 \pm 0.94$ & $71.05 \pm 2.43$ \\
 & LoRA-32 & 8.72\% & $84.95 \pm 1.82$ & $91.40 \pm 1.16$ & $73.10 \pm 2.73$ \\
 & LoRA-64 & 12.68\% & $84.07 \pm 1.41$ & $90.84 \pm 0.90$ & $71.61 \pm 2.10$ \\
 & Surgical & 28.57\% & $82.42 \pm 3.31$ & $89.70 \pm 2.23$ & $68.77 \pm 5.73$ \\
 & Full-FT & 100.00\% & $85.94 \pm 2.08$ & $92.04 \pm 1.30$ & $74.81 \pm 3.35$ \\
 & \textbf{SPECTRA} & 6.49\% & $\mathbf{87.19 \pm 0.82}$ & $\mathbf{92.83 \pm 0.52}$ & $\mathbf{76.85 \pm 1.48}$ \\
\midrule
\multirow{7}{=}{FireScars} & LP & 4.75\% & $62.61 \pm 1.00$ & $76.82 \pm 0.86$ & $68.66 \pm 1.06$ \\
 & LoRA-16 & 6.73\% & $64.30 \pm 0.68$ & $78.16 \pm 0.55$ & $69.12 \pm 1.16$ \\
 & LoRA-32 & 8.72\% & $64.97 \pm 0.30$ & $78.67 \pm 0.28$ & $69.43 \pm 1.23$ \\
 & LoRA-64 & 12.68\% & $66.79 \pm 0.63$ & $80.01 \pm 0.44$ & $71.03 \pm 1.42$ \\
 & Surgical & 28.57\% & $65.15 \pm 2.02$ & $78.78 \pm 1.52$ & $69.75 \pm 1.52$ \\
 & Full-FT & 100.00\% & $71.61 \pm 0.71$ & $83.37 \pm 0.50$ & $76.23 \pm 0.80$ \\
 & \textbf{SPECTRA} & 7.73\% & $\mathbf{68.37 \pm 1.53}$ & $\mathbf{81.10 \pm 1.10}$ & $\mathbf{73.33 \pm 1.35}$ \\
\midrule
\multirow{7}{=}{Landslide4Sense} & LP & 4.75\% & $76.00 \pm 0.30$ & $84.52 \pm 0.25$ & $53.76 \pm 0.57$ \\
 & LoRA-16 & 6.73\% & $77.46 \pm 0.04$ & $85.70 \pm 0.03$ & $56.43 \pm 0.07$ \\
 & LoRA-32 & 8.72\% & $77.56 \pm 0.16$ & $85.77 \pm 0.14$ & $56.61 \pm 0.35$ \\
 & LoRA-64 & 12.68\% & $77.58 \pm 0.13$ & $85.78 \pm 0.11$ & $56.62 \pm 0.28$ \\
 & Surgical & 28.57\% & $77.37 \pm 0.25$ & $85.62 \pm 0.20$ & $56.23 \pm 0.46$ \\
 & Full-FT & 100.00\% & $77.26 \pm 0.13$ & $85.53 \pm 0.11$ & $56.00 \pm 0.27$ \\
 & \textbf{SPECTRA} & 6.36\% & $\mathbf{78.26 \pm 0.13}$ & $\mathbf{86.33 \pm 0.11}$ & $\mathbf{57.94 \pm 0.29}$ \\
\midrule
\multirow{7}{=}{SA Crop Type} & LP & 4.75\% & $30.18 \pm 0.06$ & $43.28 \pm 0.17$ & $26.69 \pm 0.66$ \\
 & LoRA-16 & 6.73\% & $32.42 \pm 0.35$ & $45.71 \pm 0.21$ & $29.06 \pm 2.46$ \\
 & LoRA-32 & 8.72\% & $32.77 \pm 0.41$ & $46.07 \pm 0.39$ & $30.29 \pm 0.57$ \\
 & LoRA-64 & 12.68\% & $32.69 \pm 0.14$ & $46.17 \pm 0.20$ & $29.90 \pm 2.14$ \\
 & Surgical & 28.57\% & $31.20 \pm 0.63$ & $44.28 \pm 0.81$ & $28.85 \pm 1.84$ \\
 & Full-FT & 100.00\% & $33.30 \pm 1.12$ & $46.54 \pm 1.54$ & $33.04 \pm 2.38$ \\
 & \textbf{SPECTRA} & 7.73\% & $\mathbf{33.39 \pm 0.47}$ & $\mathbf{46.78 \pm 0.49}$ & $\mathbf{28.51 \pm 0.51}$ \\
  \bottomrule
  \end{tabular}
\end{table*}

\section{Experiments}

\subsection{Experimental Setup}

\paragraph{GeoFM backbones and datasets.}
As listed in Tables~\ref{tab:geofm_backbones} and~\ref{tab:datasets} respectively, we evaluate three GeoFM backbones: Prithvi-EO-2.0 600M \cite{roy2024prithvi}, ScaleMAE \cite{reed2023scalemae} and SatMAE \cite{cong2022satmae}, and four downstream datasets: Sen1Floods11 \cite{bonafilia2020sen1floods11}, FireScars, Landslide4Sense \cite{ghorbanzadeh2022landslide4sense}, and GEO-Bench South America crop-type segmentation \cite{lacoste2023geobench}. These combinations cover binary and multiclass segmentation and expose different degrees of spectral mismatch with each backbone.

\paragraph{Baselines.}
We compare \method{} with four fine-tuning baselines. 
\emph{Linear probing} (LP) freezes the pretrained patch embedding and encoder and
trains only the task-specific decoder, following the common practice of
evaluating a frozen representation with a trainable readout~\cite{kornblith2019better}.
\emph{Uniform LoRA} freezes the pretrained backbone and trains LoRA adapters with
a fixed rank $r\in\{8,16,32,64\}$ in every adapted transformer block, together
with the decoder~\cite{hu2022lora}. For example, the method LoRA-32 applies a uniform rank of 32 throughout the pretrained encoder.
\emph{Surgical tuning} unfreezes the encoder stage selected by auto-RGN, while keeping the remaining stages frozen and training the decoder~\cite{lee2023surgical}. 
\emph{Full fine-tuning} (\textbf{Full-FT}) trains the full pretrained backbone and the
decoder, and \textbf{is treated as a high-cost reference rather than the main
competitor}. 
Unless otherwise specified, all baselines use \bandsel{} as the input setting.
For component studies of \bre{}, we use LoRA-32 with \bandsel{} as the matched
reference unless otherwise stated. To test whether \bre{} generalizes beyond
one fine-tuning policy, we also compare each baseline policy under its original
\bandsel{} input, a direct all-band MLP projector, and the same policy augmented
with \bre{}.

\paragraph{Decoder, splits, and metrics.}
All runs use UPerNet \cite{xiao2018upernet} decoder so that the comparison focuses on input adaptation and encoder adaptation rather than decoder choice. We use seeds 42, 43, and 44, with fixed matched settings inside each backbone-dataset cell. We report test macro \miou{}, foreground IoU~\cite{long2015fully}, and macro F1~\cite{vanrijsbergen1979information,sokolova2009systematic} for all segmentation tasks. All results are evaluated on the held-out test set using the best validation checkpoint selected during 50-epoch training. Parameter-efficiency comparisons report trainable parameters and trainable percentage relative to the full backbone-plus-decoder model defined as Equation \ref{eq:percent}.

\paragraph{Other training hyperparameters.}
For each GeoFM--dataset pair, we keep the loss function, learning rates, decoder
architecture, and data split. The training objective combines cross-entropy~\cite{goodfellow2016deep} and Dice loss~\cite{milletari2016vnet} for segmentation with class imbalance, with
their relative weights adjusted by dynamic weight averaging
(DWA)~\cite{liu2019end}. Detailed hyperparameters for all experiments are
reported in our \href{https://github.com/big-data-lab-umbc/SPECTRA/blob/main/appendix_experiment_hyperparameters.md}{\texttt{GitHub repository}}. All
experiments were run on a single NVIDIA L40S GPU with 48 GB of VRAM.

% \paragraph{Baselines}.
%  LP trains only the decoder. Uniform LoRA trains LoRA updates with ranks 8, 16, 32, or 64. Last-stage tuning unfreezes the last encoder stage and the decoder. Surgical tuning unfreezes the LogME-selected stage and the decoder. Full-FT trains the full backbone and decoder. For component studies, LoRA-32 with \bandsel{} is the matched reference for \bre{}. Full fine-tuning is treated as a high-cost reference with upper-bound performance rather than the main competitor. All baselines use \bandsel{} as the input setting.

\subsection{Main Comparison}

% Table~\ref{tab:main_comparison} reports the full main comparison. 

% The adaptation policies are fixed as follows. LP trains only the decoder and does not attach LoRA modules. Uniform LoRA uses the same rank at every adapted encoder stage, so LoRA-16, LoRA-32, and LoRA-64 use $r=16$, $r=32$, and $r=64$, respectively.

% The main comparison supports the parameter-efficiency motivation but also shows why claims must be scoped. On Prithvi with Sen1Floods11, \method{} with \bre{} and ST-LoRA reaches $86.39 \pm 0.85$ macro \miou{} while updating only $4.06\%$ of the model. This exceeds the included LoRA and full fine-tuning references while using far less trainable capacity than full fine-tuning. On ScaleMAE with Sen1Floods11, \method{} with \bre{} and ST-LoRA Transfer reaches $85.45 \pm 2.08$ macro \miou{} with $7.23\%$ trainable parameters, above the matched \bandsel{} LoRA-64 and full-FT references.

% These results do not imply that \method{} dominates every baseline across every backbone-dataset pair. The stable claim is that \method{} provides a practical way to combine extra-band use with lower-budget stage-wise LoRA, and that the completed stage-wise rows show competitive accuracy at substantially lower adaptation cost than high-capacity baselines.

Full comparison across all backbones and datasets is in Table~\ref{tab:main_comparison}, and it is organized by GeoFM backbone and dataset. Each block compares different adaptation methods under the same dataset and GeoFM. 

Overall, \method{} achieves strong accuracy with only $3.29$--$7.73\%$ trainable parameters. It is consistently competitive with uniform LoRA and often improves
over it with a smaller adaptation budget. The gains are especially clear for
ScaleMAE and SatMAE, where \method{} is the best or near-best method on most
dataset--metric pairs.

The largest improvements occur on ScaleMAE. On Sen1Floods11, \method{} improves
macro \miou{} from $77.17$ with LoRA-64 and $75.41$ with Full-FT to $85.45$.
On FireScars, it improves macro \miou{} from $69.03$ with LoRA-64 and $72.30$
with Full-FT to $82.05$. \method{} also exceeds Full-FT on several other settings,
including Prithvi-EO-2.0 on Sen1Floods11 and Landslide4Sense, and SatMAE on
Sen1Floods11, Landslide4Sense, and SA Crop Type in macro \miou{}.

\method{} does not improve every case. For Prithvi-EO-2.0 on FireScars, uniform
LoRA-64 and Full-FT remain stronger than \method{}. The reason might be that the pretrained bands and finetuning bands are natively matched, and the backbone is already strong for FireScars. To sum up, these results indicate that \method{} is not simply a universal
replacement for high-capacity encoder adaptation, but a strong low-cost alternative when spectral adaptation is beneficial.

\begin{figure}[t]
  \centering
  \includegraphics[width=\columnwidth]{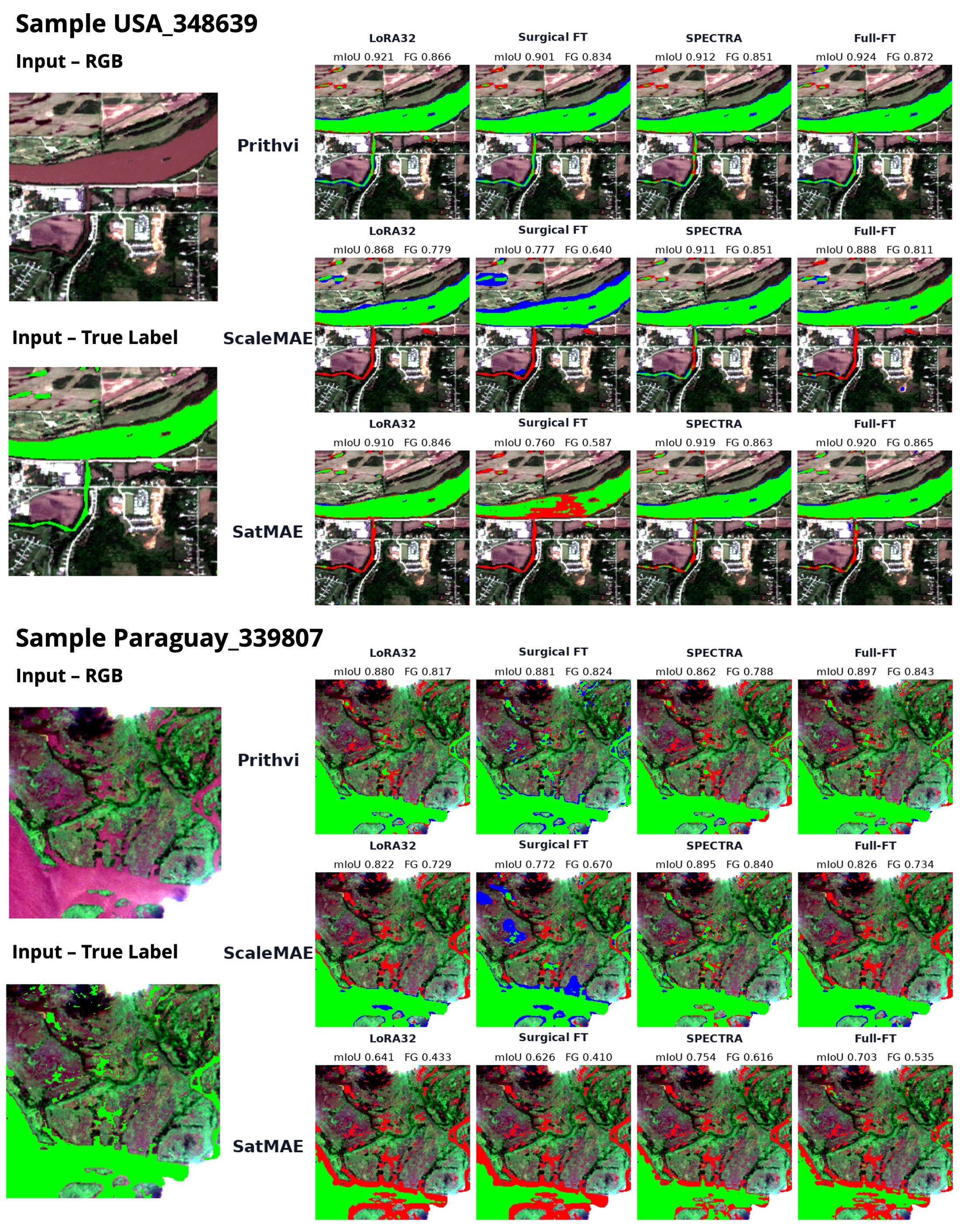}
  \Description{Qualitative segmentation examples comparing LoRA-32 band selection with SPECTRA across three representative binary segmentation examples. Each example shows RGB input, true label, and prediction overlays for ScaleMAE, SatMAE, and Prithvi. Green marks true positives, blue marks false positives, and red marks false negatives.}
  \caption{Visualization of predictions for LoRA-32 \bandsel{}, surgical tuning, \method{} and full fine-tuning on representative binary segmentation examples. Green, blue, and red denote true positives, false positives, and false negatives, respectively.}
  \label{fig:qualitative_predictions}
\end{figure}

Besides the quantitative comparison, Figure~\ref{fig:qualitative_predictions} provides visualizations of examples. The ScaleMAE predictions show the clearest visual improvement, with \method{} recovering foreground regions missed by the LoRA-32 \bandsel{} baseline. And the \method{} visualizations of ScaleMAE are even comparable to Prithvi, which natively learns strong features for the floods dataset. The compared baseline method Sugical FT is strong for Prithvi, but does not make good predictions when used with ScaleMAE and SatMAE. Overall speaking, SPECTRA not only enhances quantitative metrics but also makes qualitative improvements for GeoFMs that have large band mismatching with downstream tasks.

\begin{figure}[t]
  \centering
  \includegraphics[width=\columnwidth]{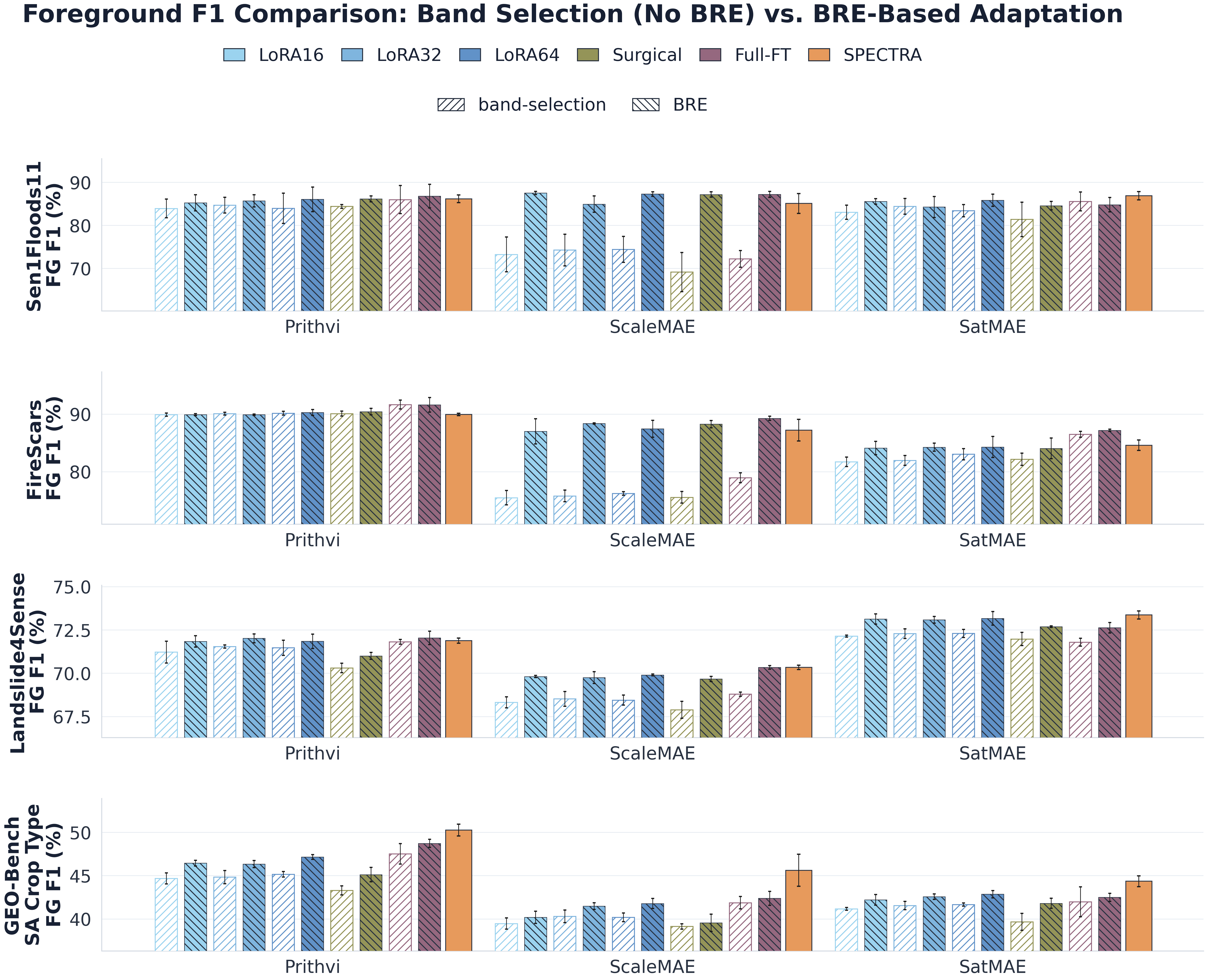}
  \Description{Grouped bar charts of foreground F1 across datasets, GeoFM backbones, and fine-tuning baselines. Hatched bars use band selection without BRE, solid bars add BRE to the same baseline, and orange bars show the full SPECTRA pipeline.}
  % \caption{Comparison of band-selection and BRE-based adaptation applied to different fine-tuning methods. Evaluated by Foreground (FG) F1 score across different GeoFMs and datasets. For each fine-tuning method (illustrated by the same color), . Orange bars show the full proposed \method{} pipeline with both \bre{} and \starplanner{}.}
    \caption{Comparison of \bandsel{} and BRE-based adaptation across fine-tuning methods, evaluated by foreground (FG) F1 on multiple GeoFMs and datasets. Bars of the same color represent the same fine-tuning method, while the two fill patterns distinguish \bandsel{} from BRE. Orange bars show the full \method{} pipeline with \bre{} and \starplanner{}.}
  \label{fig:bre_grouped_bar}
\end{figure}

\begin{table}[t]
  \centering
  \small
  \setlength{\tabcolsep}{2.0pt}
  \renewcommand{\arraystretch}{0.95}
  \caption{Comparison between MLP projection and \bre{}. Values are average test macro \miou{} (\%) over $n$ matched GeoFM--dataset pairs from three GeoFMs and four datasets. This table excludes \starplanner{} and compares only each FT method with \bandsel{}, +MLP, and +\bre{} input adaptation. Deltas are relative to the matched \bandsel{} baseline.}
  \label{tab:mlp_bre_adapter_summary}
  \begin{tabular}{@{}lrrrrrr@{}}
    \toprule
    Scope & $n$ & \bandsel{} & +MLP & +\bre{} & $\Delta$MLP & $\Delta$\bre{} \\
    \midrule
    LoRA-16 & 12 & 65.89 & 64.26 & \textbf{68.95} & $-1.63$ & $\mathbf{+3.06}$ \\
    LoRA-32 & 12 & 66.32 & 63.59 & \textbf{68.90} & $-2.73$ & $\mathbf{+2.59}$ \\
    LoRA-64 & 12 & 66.45 & 63.84 & \textbf{69.30} & $-2.61$ & $\mathbf{+2.85}$ \\
    Surgical & 12 & 65.37 & 63.20 & \textbf{68.84} & $-2.16$ & $\mathbf{+3.48}$ \\
    Full-FT & 12 & 67.77 & 65.49 & \textbf{70.20} & $-2.28$ & $\mathbf{+2.43}$ \\
    \bottomrule
  \end{tabular}
\end{table}

\begin{figure}[t]
  \centering
  \includegraphics[width=0.96\columnwidth]{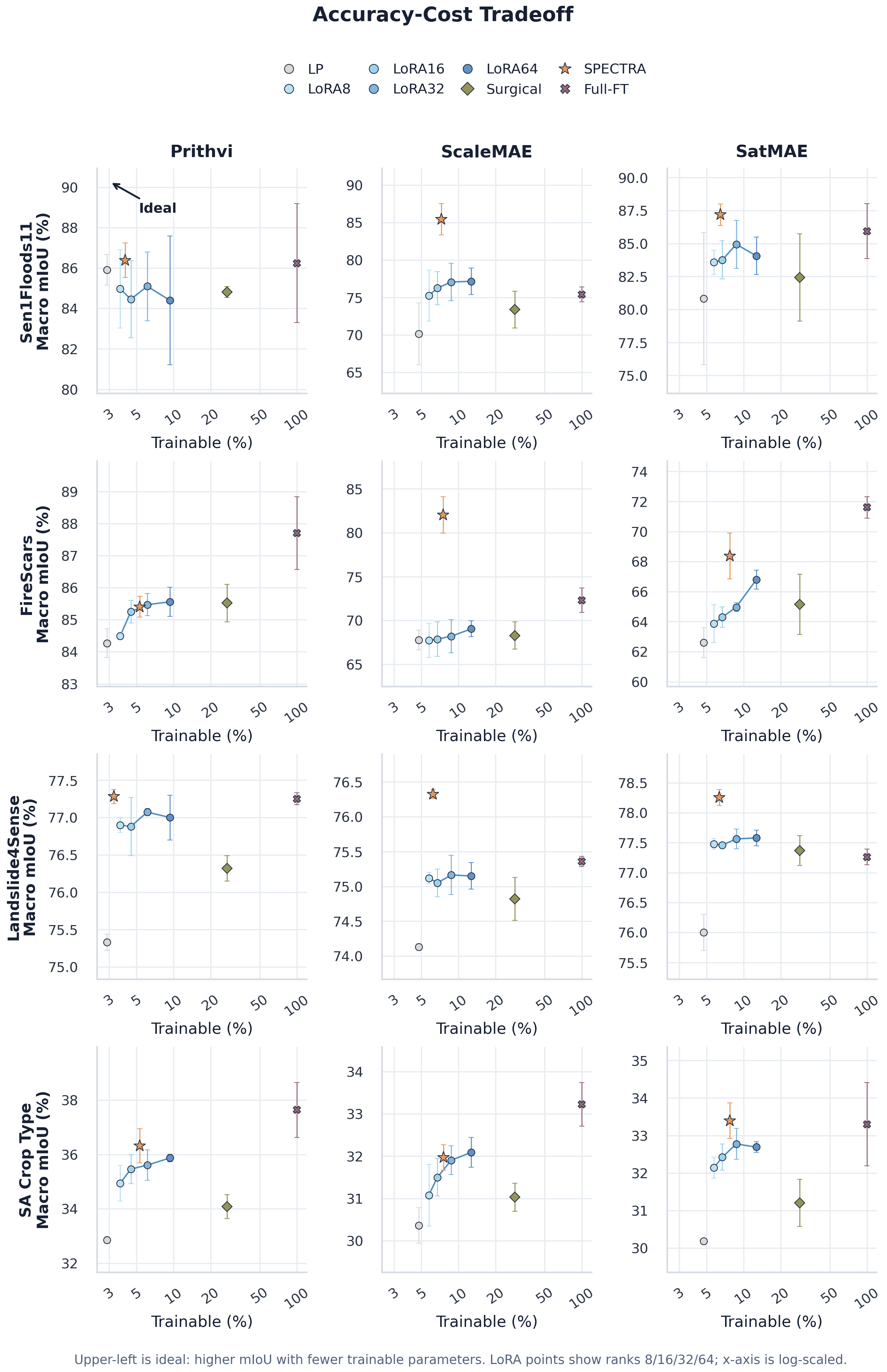}
  \Description{Single-column accuracy-cost plots across GeoFMs and segmentation tasks. Each panel plots trainable parameter percentage on the x-axis and macro mIoU on the y-axis, with separate marks for LP, the LoRA rank curve, surgical fine-tuning, SPECTRA, and full fine-tuning.}
  \caption{Accuracy-cost tradeoff curves across GeoFMs and datasets. Markers compare LP, uniform LoRA ranks, surgical tuning, \method{}, and full fine-tuning.}
  \label{fig:accuracy_cost_curve}
\end{figure}

\subsection{Does Extra-Band Embedding Help?}

Table~\ref{tab:mlp_bre_adapter_summary} tests whether using all available bands is beneficial and how a direct learned MLP-based projection performs compared to BRE.
The table aggregates matched results over three GeoFMs, four datasets, and five baseline fine-tuning policies. The \bandsel{} column is the same selected-band baseline summarized in Table~\ref{tab:main_comparison}, the MLP column inserts a three-layer all-band projector before
the pretrained patch embedding, and the \bre{} column replaces only the input
adapter while keeping the same baseline fine-tuning policy. Overall, \bre{}
raises mean macro \miou{} from $66.36$ to $69.24$ ($+2.88$ points), while MLP
projection decreases it to $64.08$ ($-2.28$ points). Appendix
Tables~\ref{tab:appendix_full_mlp_prithvi}--\ref{tab:appendix_full_mlp_satmae}
report the full matched baseline-versus-MLP rows behind the MLP column.

This comparison shows that extra bands are useful, but that the input adapter
must preserve the pretrained spectral interface. A direct MLP uses all channels,
but it changes the input distribution before the pretrained patch embedding and
mixes selected and extra bands from a randomly initialized projection. The model
therefore has to relearn a spectral tokenizer while also adapting to the
downstream task. \bre{} avoids this failure mode: it starts exactly as
\bandsel{} through the selected-band anchor and adds a zero-initialized
all-band residual path. Thus, extra bands can improve the compatible input when
they are useful, while the model begins from the stable pretrained interface.

The broader component comparison in Figure~\ref{fig:bre_grouped_bar} shows that
this pattern is not limited to one dataset, one GeoFM, or one fine-tuning
policy. For each policy, the hatched bar is the original \bandsel{} input and
the solid bar adds \bre{} while keeping the same encoder-adaptation policy.
Across the 60 paired baseline-policy cells, \bre{} improves
foreground/no-background F1 in 56 cells, with an average gain of $+3.08$ F1
points. The corresponding foreground IoU gain averages $+4.15$
points. The gains are positive on average for all three backbones, all four
datasets, and all five baseline policies, including LoRA-16/32/64, Surgical,
and Full-FT.

The largest gains occur when fixed band selection discards task-relevant
spectral information. For example, Figure~\ref{fig:bre_grouped_bar} shows large
improvements for ScaleMAE on Sen1Floods11 and FireScars, where the RGB-limited
pretrained interface forces \bandsel{} to ignore many multispectral target
bands. Boundary cases remain important: Prithvi on FireScars is nearly native
to the HLS six-band interface, so \bre{} provides little benefit for that pair,
and a small number of cells remain negative. These cases support a conditional
interpretation: \bre{} is most useful when extra observed bands contain
information that the selected-band baseline discards, and it is a more
appropriate all-band adaptation mechanism than an unconstrained MLP projector.

\subsection{Does \starplanner{} Select Useful Stage-wise Ranks?}

Figure~\ref{fig:accuracy_cost_curve} and
Table~\ref{tab:rank_ablation_sen1floods} provide complementary views of the
rank-planning behavior. In Figure~\ref{fig:accuracy_cost_curve}, each panel
plots macro \miou{} against the percentage of trainable parameters for one
GeoFM--dataset pair. It compares LP, four LoRA experiments with uniform ranks, including [8, 16, 32, 64], surgical fine-tuning, \method{}, and full fine-tuning. Points farther left use fewer trainable parameters, and points higher are more accurate.

Table~\ref{tab:rank_ablation_sen1floods} isolates the ST-planner rank allocation
on Sen1Floods11. Rows M1--M8 are manually designed control schedules that
distribute LoRA rank across stages with different depth preferences and
budgets. They use stage-wise LoRA but \textbf{without our automatic ST-planner}. A rank vector $[r_1,r_2,r_3,r_4]$ gives the stage-wise LoRA rank assigned from shallow to deep encoder stages, and the rank budget is
$\sum_s r_s$. The row ``\method{}'' uses ST-planner to plan the rank $[r_1,r_2,r_3,r_4]$. For the final \method{} configuration in Table~\ref{tab:main_comparison}, Prithvi with Sen1Floods11 uses ST-LoRA Repair and reaches $86.39 \pm 0.85$ macro \miou{}, $+0.43$ percentage points above the uniform LoRA-32 \bre{} reference while using fewer active LoRA parameters. For ScaleMAE, the strongest automatic \bre{} schedule is ST-LoRA Transfer with rank vector $[16,32,16,16]$. It reaches $85.45 \pm 2.08$ macro \miou{}, slightly above the uniform LoRA-32 \bre{} reference. In Figure~\ref{fig:accuracy_cost_curve}, these \method{} points generally sit far left of full fine-tuning and surgical fine-tuning, while remaining competitive with or above nearby LoRA ranks. 

These results support the efficiency claim, and they also show that a larger rank
budget does not automatically lead to better performance. Manual schedules remain
useful controls when sufficient time and computing resources are available to test different
rank allocations. The advantage of \starplanner{} is that it automatically finds
an efficient stage-wise rank schedule without exhaustively searching over the
$|\mathcal{R}|^S$ possible combinations of $[r_1,r_2,r_3,r_4]$.

\begin{table}[h]
  \centering
  \small
  \setlength{\tabcolsep}{1.4pt}
  \renewcommand{\arraystretch}{0.95}
  \caption{Comparison between \starplanner{} and manually-selected ranks on Sen1Floods11.}
  \label{tab:rank_ablation_sen1floods}
  \begin{tabularx}{\columnwidth}{@{}>{\raggedright\arraybackslash}X c c c@{}}
    \toprule
    Method \& ranks & Rank budget  & Trainable params & Macro \miou{} \% \\
    \midrule

    \multicolumn{4}{@{}l}{\textbf{GeoFM: Prithvi-EO-2.0 600M}} \\
    \midrule
    M1 [4,4,8,16]
      & 32 & 23.94M & $86.58 \pm 0.71$   \\
    M2 [16,8,4,4]
      & 32 & 23.94M & $84.98 \pm 2.37$  \\
    M3 [8,16,16,8]
      & 48 & 26.57M & $85.85 \pm 2.13$  \\
    M4 [4,8,16,32]
      & 60 & 28.53M & $85.85 \pm 1.47$  \\
    M5 [32,16,16,8]
      & 72 & 30.50M & $85.02 \pm 2.39$ \\
    M6 [8,16,24,32]
      & 80 & 31.81M & $85.22 \pm 1.89$  \\
    M7 [4,8,16,64]
      & 92 & 33.77M & $86.05 \pm 1.05$ \\
    M8 [64,32,16,8]
      & 120 & 38.36M & $85.15 \pm 2.89$  \\
    \starplanner{} [8,16,16,8]
      & 48 & 26.35M & $85.88 \pm 1.24$  \\

    \midrule
    \multicolumn{4}{@{}l}{\textbf{GeoFM: ScaleMAE}} \\
    \midrule
   M1 [4,4,8,16]
      & 32 & 18.23M & $85.27 \pm 1.77$  \\
   M2 [16,8,4,4]
      & 32 & 18.23M & $84.98 \pm 1.82$  \\
   M3 [8,16,16,8]
      & 48 & 19.80M & $84.63 \pm 1.75$ \\
   M4 [4,8,16,32]
      & 60 & 20.98M & $85.15 \pm 1.83$  \\
   M5 [32,16,16,8]
      & 72 & 22.16M & $84.99 \pm 1.67$  \\
   M6 [8,16,24,32]
      & 80 & 22.95M & $85.25 \pm 1.68$  \\
   M7 [4,8,16,64]
      & 92 & 24.13M & $85.12 \pm 1.89$  \\
   M8 [64,32,16,8]
      & 120 & 26.88M & $85.01 \pm 1.59$  \\
   \starplanner{} [16,32,16,16]
      & 80 & 22.95M & $85.29 \pm 2.17$ \\
    \bottomrule
  \end{tabularx}
\end{table}

\section{Discussion and Limitations}

\textit{When should extra-band adapters be used?}
The practical rule from the experiments is conditional. If spectral mismatch is low and \bandsel{} is already strong, the safest default may be to keep \bandsel{}. The Prithvi--FireScars result illustrates this boundary condition: the pretrained and fine-tuning inputs share the same six HLS spectral bands, so there are no discarded channels for \bre{} to recover, and the grouped-bar results show no performance enhancement. If many target bands are discarded or the target sensor contains task-relevant channels outside the selected subset, \bre{} should be tested under matched training settings. The largest observed gains are consistent with this rule, especially for ScaleMAE on multispectral downstream datasets.

% \textit{Why preserve the pretrained patch embedding?}
% The patch embedding is part of the pretrained representation. Replacing it forces the downstream model to learn a new spectral tokenizer and a task decoder at the same time. \bre{} avoids this failure mode by starting exactly from \bandsel{} and allowing a zero-initialized residual correction to grow only if it improves the supervised objective. This design makes the extra-band path conservative: it can use additional channels without discarding the pretrained interface.

\textit{Limitations.}
The current study uses constrained hyperparameters and a common UPerNet decoder to control variables across backbones and datasets. This is appropriate for method diagnosis, but it is not a claim of published state-of-the-art performance on every dataset. BRE gains are not universal, and their magnitude can change by backbone and task. The \starplanner{} evidence is strongest on the completed Sen1Floods11 rank-planner runs; broader rank-planner validation across all downstream datasets remains necessary before claiming universal parameter-efficiency gains. Appendix~\ref{app:published_prithvi_comparison} further compares our results with published Prithvi-EO-2.0 task-specific fine-tuning results to clarify how these controlled adaptation results should be interpreted for domain-science and product use.

\flushbottom
\section{Conclusion}

We introduced \method{}, a parameter-efficient fine-tuning framework for cross-sensor GeoFM fine-tuning under spectral mismatch. The proposed \bre{} routes all available target bands into a pretrained-compatible virtual input, while \starplanner{} allocates LoRA rank across encoder stages using transferability diagnostics. Overall speaking, \method{} improves both performance and efficiency on different GeoFMs and datasets. The results reveal that extra-band embedding helps most when \bandsel{} discards useful channels, and stage-wise rank planning can reduce LoRA budget while preserving competitive accuracy.

\bibliographystyle{ACM-Reference-Format}
\bibliography{references}

@inproceedings{you2021logme,
  title = {{LogME}: Practical Assessment of Pre-trained Models for Transfer Learning},
  author = {You, Kaichao and Liu, Yong and Wang, Jianmin and Long, Mingsheng},
  booktitle = {Proceedings of the 38th International Conference on Machine Learning},
  pages = {12133--12143},
  year = {2021},
  volume = {139},
  series = {Proceedings of Machine Learning Research},
  publisher = {PMLR},
  url = {https://proceedings.mlr.press/v139/you21b.html}
}

@inproceedings{nguyen2020leep,
  title = {{LEEP}: A New Measure to Evaluate Transferability of Learned Representations},
  author = {Nguyen, Cuong and Hassner, Tal and Seeger, Matthias and Archambeau, Cedric},
  booktitle = {Proceedings of the 37th International Conference on Machine Learning},
  year = {2020},
  series = {Proceedings of Machine Learning Research}
}

@inproceedings{xiao2018upernet,
  title = {Unified Perceptual Parsing for Scene Understanding},
  author = {Xiao, Tete and Liu, Yingcheng and Zhou, Bolei and Jiang, Yuning and Sun, Jian},
  booktitle = {Proceedings of the European Conference on Computer Vision},
  pages = {418--434},
  year = {2018}
}

@inproceedings{cong2022satmae,
  title = {{SatMAE}: Pre-training Transformers for Temporal and Multi-Spectral Satellite Imagery},
  author = {Cong, Yezhen and Khanna, Samar and Meng, Chenlin and Liu, Patrick and Rozi, Erik and He, Yutong and Burke, Marshall and Lobell, David B. and Ermon, Stefano},
  booktitle = {Advances in Neural Information Processing Systems},
  volume = {35},
  pages = {197--211},
  year = {2022}
}

@inproceedings{reed2023scalemae,
  title = {Scale-{MAE}: A Scale-Aware Masked Autoencoder for Multiscale Geospatial Representation Learning},
  author = {Reed, Colorado J. and Gupta, Ritwik and Li, Shufan and Brockman, Sarah and Funk, Christopher and Clipp, Brian and Keutzer, Kurt and Candido, Salvatore and Uyttendaele, Matt and Darrell, Trevor},
  booktitle = {Proceedings of the IEEE/CVF International Conference on Computer Vision},
  pages = {4088--4099},
  month = {October},
  year = {2023}
}

@misc{jakubik2023foundation,
  title = {Foundation Models for Generalist Geospatial Artificial Intelligence},
  author = {Jakubik, Johannes and Roy, Sujit and Phillips, Christopher E. and Fraccaro, Paolo and Godwin, Denys and Zadrozny, Bianca and Szwarcman, Daniela and Gomes, Carlos and Nyirjesy, Gabby and Edwards, Blair and others},
  year = {2023},
  eprint = {2310.18660},
  archivePrefix = {arXiv},
  primaryClass = {cs.CV},
  url = {https://arxiv.org/abs/2310.18660}
}

@article{roy2024prithvi,
  title = {Prithvi-{EO}-2.0: A Versatile Multi-Temporal Foundation Model for Earth Observation Applications},
  author = {Szwarcman, Daniela and Roy, Sujit and Fraccaro, Paolo and Gislason, Orsteinn Eli and Blumenstiel, Benedikt and Ghosal, Rinki and de Oliveira, Pedro Henrique and de Sousa Almeida, Joao Lucas and Sedona, Rocco and Kang, Yanghui and Chakraborty, Srija and Wang, Sizhe and Gomes, Carlos and Kumar, Ankur and Gaur, Vishal and Truong, Myscon and Godwin, Denys and Khallaghi, Sam and Lee, Hyunho and Hsu, Chia Yu and Asanjan, Ata Akbari and Mujeci, Besart and Shidham, Disha and Balogun, Rufai Omowunmi and Kolluru, Venkatesh and Keenan, Trevor and Arevalo, Paulo and Li, Wenwen and Alemohammad, Hamed and Olofsson, Pontus and Mayer, Timothy and Hain, Christopher and Kennedy, Robert and Zadrozny, Bianca and Bell, David and Cavallaro, Gabriele and Watson, Campbell and Maskey, Manil and Ramachandran, Rahul and Moreno, Juan Bernabe},
  journal = {IEEE Transactions on Geoscience and Remote Sensing},
  year = {2025},
  doi = {10.1109/TGRS.2025.3642610}
}

@inproceedings{lacoste2023geobench,
  title = {{GEO-Bench}: Toward Foundation Models for Earth Monitoring},
  author = {Lacoste, Alexandre and Lehmann, Nils and Rodriguez, Pau and Sherwin, Evan and Kerner, Hannah and L{"u}tjens, Bj{"o}rn and Irvin, Jeremy and Dao, David and Alemohammad, Hamed and Drouin, Alexandre and Gunturkun, Mehmet and Huang, Gabriel and Vazquez, David and Newman, Dava and Bengio, Yoshua and Ermon, Stefano and Zhu, Xiaoxiang},
  booktitle = {Advances in Neural Information Processing Systems},
  volume = {36},
  year = {2023}
}

@inproceedings{bonafilia2020sen1floods11,
  title = {Sen1Floods11: A Georeferenced Dataset to Train and Test Deep Learning Flood Algorithms for Sentinel-1},
  author = {Bonafilia, Derrick and Tellman, Beth and Anderson, Tyler and Issenberg, Erica},
  booktitle = {Proceedings of the IEEE/CVF Conference on Computer Vision and Pattern Recognition Workshops},
  pages = {835--845},
  year = {2020}
}

@article{ghorbanzadeh2022landslide4sense,
  title = {Landslide4Sense: Reference Benchmark Data and Deep Learning Models for Landslide Detection},
  author = {Ghorbanzadeh, Omid and Xu, Yonghao and Ghamisi, Pedram and Kopp, Michael and Kreil, David},
  journal = {IEEE Transactions on Geoscience and Remote Sensing},
  volume = {60},
  pages = {1--17},
  year = {2022},
  doi = {10.1109/TGRS.2022.3215209}
}

@article{mai2024opportunities,
  title = {On the Opportunities and Challenges of Foundation Models for {GeoAI}},
  author = {Mai, Gengchen and Huang, Weiming and Sun, Jin and Song, Shenzhe and Mishra, Deepak and Liu, Ninghao and Gao, Song and Liu, Tianming and Cong, Gao and Hu, Yingjie and Cundy, Chris and Li, Ziyuan and Zhu, Rui and Lao, Ni},
  journal = {ACM Transactions on Spatial Algorithms and Systems},
  volume = {10},
  number = {2},
  articleno = {11},
  pages = {1--46},
  year = {2024},
  doi = {10.1145/3653070}
}

@article{marsocci2026pangaea,
  title = {{PANGAEA}: Assessing Geospatial Foundation Models Capabilities through a Global and Inclusive Benchmark},
  author = {Marsocci, Valerio and Jia, Yuru and Le Bellier, Georges and Kerekes, David and Zeng, Liang and Hafner, Sebastian and Gerard, Sebastian and Brune, Eric and Yadav, Ritu and Shibli, Ali and Fang, Heng and Ban, Yifang and Vergauwen, Maarten and Audebert, Nicolas and Nascetti, Andrea},
  journal = {IEEE Geoscience and Remote Sensing Magazine},
  volume = {14},
  number = {1},
  pages = {245--285},
  year = {2026},
  doi = {10.1109/MGRS.2025.3628194}
}

@inproceedings{dosovitskiy2021image,
  title = {An Image is Worth 16x16 Words: Transformers for Image Recognition at Scale},
  author = {Dosovitskiy, Alexey and Beyer, Lucas and Kolesnikov, Alexander and Weissenborn, Dirk and Zhai, Xiaohua and Unterthiner, Thomas and Dehghani, Mostafa and Minderer, Matthias and Heigold, Georg and Gelly, Sylvain and Uszkoreit, Jakob and Houlsby, Neil},
  booktitle = {International Conference on Learning Representations},
  year = {2021}
}

@inproceedings{he2022masked,
  title = {Masked Autoencoders Are Scalable Vision Learners},
  author = {He, Kaiming and Chen, Xinlei and Xie, Saining and Li, Yanghao and Doll{\'a}r, Piotr and Girshick, Ross},
  booktitle = {Proceedings of the IEEE/CVF Conference on Computer Vision and Pattern Recognition},
  pages = {16000--16009},
  year = {2022}
}

@inproceedings{houlsby2019parameter,
  title = {Parameter-Efficient Transfer Learning for {NLP}},
  author = {Houlsby, Neil and Giurgiu, Andrei and Jastrzebski, Stanislaw and Morrone, Bruna and de Laroussilhe, Quentin and Gesmundo, Andrea and Attariyan, Mona and Gelly, Sylvain},
  booktitle = {Proceedings of the 36th International Conference on Machine Learning},
  series = {Proceedings of Machine Learning Research},
  volume = {97},
  pages = {2790--2799},
  publisher = {PMLR},
  year = {2019}
}

@article{dong2024upetu,
  title = {{UPetu}: A Unified Parameter-Efficient Fine-Tuning Framework for Remote Sensing Foundation Model},
  author = {Dong, Zhe and Gu, Yanfeng and Liu, Tianzhu},
  journal = {IEEE Transactions on Geoscience and Remote Sensing},
  volume = {62},
  pages = {1--13},
  year = {2024},
  doi = {10.1109/TGRS.2024.3382734}
}

@inproceedings{marti2025fine,
  title={Fine-tune smarter, not harder: Parameter-efficient fine-tuning for geospatial foundation models},
  author={Marti Escofet, Francesc and Blumenstiel, Benedikt and Scheibenreif, Linus and Fraccaro, Paolo and Schindler, Konrad},
  booktitle={Joint European Conference on Machine Learning and Knowledge Discovery in Databases},
  pages={516--532},
  year={2025},
  organization={Springer}
}

@inproceedings{lee2023surgical,
  title = {Surgical Fine-Tuning Improves Adaptation to Distribution Shifts},
  author = {Lee, Yoonho and Chen, Annie S. and Tajwar, Fahim and Kumar, Ananya and Yao, Huaxiu and Liang, Percy and Finn, Chelsea},
  booktitle = {International Conference on Learning Representations},
  year = {2023}
}

@misc{xiong2024dofa,
  title = {Neural Plasticity-Inspired Multimodal Foundation Model for Earth Observation},
  author = {Xiong, Zhitong and Wang, Yi and Zhang, Fahong and Stewart, Adam J. and Hanna, Jo{"e}lle and Borth, Damian and Papoutsis, Ioannis and Le Saux, Bertrand and Camps-Valls, Gustau and Zhu, Xiao Xiang},
  year = {2024},
  eprint = {2403.15356},
  archivePrefix = {arXiv},
  primaryClass = {cs.CV},
  url = {https://arxiv.org/abs/2403.15356}
}

@inproceedings{han2024msgfm,
  title = {Bridging Remote Sensors with Multisensor Geospatial Foundation Models},
  author = {Han, Boran and Zhang, Shuai and Shi, Xingjian and Reichstein, Markus},
  booktitle = {Proceedings of the IEEE/CVF Conference on Computer Vision and Pattern Recognition},
  pages = {27852--27862},
  year = {2024},
  doi = {10.1109/CVPR52733.2024.02631}
}

@inproceedings{astruc2025anysat,
  title = {{AnySat}: One Earth Observation Model for Many Resolutions, Scales, and Modalities},
  author = {Astruc, Guillaume and Gonthier, Nicolas and Mallet, Cl{\'e}ment and Landrieu, Lo{\"i}c},
  booktitle = {Proceedings of the IEEE/CVF Conference on Computer Vision and Pattern Recognition},
  pages = {19530--19540},
  year = {2025},
  doi = {10.1109/CVPR52734.2025.01819}
}

@article{dosovitskiy2020image,
  title={An image is worth 16x16 words: Transformers for image recognition at scale},
  author={Dosovitskiy, Alexey and Beyer, Lucas and Kolesnikov, Alexander and Weissenborn, Dirk and Zhai, Xiaohua and Unterthiner, Thomas and Dehghani, Mostafa and Minderer, Matthias and Heigold, Georg and Gelly, Sylvain and others},
  journal={arXiv preprint arXiv:2010.11929},
  year={2020}
}

@inproceedings{liu2019end,
  title={End-to-end multi-task learning with attention},
  author={Liu, Shikun and Johns, Edward and Davison, Andrew J},
  booktitle={Proceedings of the IEEE/CVF conference on computer vision and pattern recognition},
  pages={1871--1880},
  year={2019}
}

@inproceedings{long2015fully,
  title     = {Fully Convolutional Networks for Semantic Segmentation},
  author    = {Long, Jonathan and Shelhamer, Evan and Darrell, Trevor},
  booktitle = {Proceedings of the IEEE Conference on Computer Vision and Pattern Recognition},
  pages     = {3431--3440},
  year      = {2015}
}

@book{vanrijsbergen1979information,
  title     = {Information Retrieval},
  author    = {van Rijsbergen, Cornelis Joost},
  edition   = {2},
  publisher = {Butterworth-Heinemann},
  year      = {1979}
}

@article{sokolova2009systematic,
  title   = {A systematic analysis of performance measures for classification tasks},
  author  = {Sokolova, Marina and Lapalme, Guy},
  journal = {Information Processing \& Management},
  volume  = {45},
  number  = {4},
  pages   = {427--437},
  year    = {2009}
}

@inproceedings{kornblith2019better,
  title     = {Do Better ImageNet Models Transfer Better?},
  author    = {Kornblith, Simon and Shlens, Jonathon and Le, Quoc V.},
  booktitle = {Proceedings of the IEEE/CVF Conference on Computer Vision and Pattern Recognition},
  pages     = {2661--2671},
  year      = {2019}
}

@inproceedings{hu2022lora,
  title     = {{LoRA}: Low-Rank Adaptation of Large Language Models},
  author    = {Hu, Edward J. and Shen, Yelong and Wallis, Phillip and Allen-Zhu, Zeyuan and Li, Yuanzhi and Wang, Shean and Wang, Lu and Chen, Weizhu},
  booktitle = {International Conference on Learning Representations},
  year      = {2022}
}

@book{goodfellow2016deep,
  title     = {Deep Learning},
  author    = {Goodfellow, Ian and Bengio, Yoshua and Courville, Aaron},
  publisher = {MIT Press},
  year      = {2016}
}

@inproceedings{milletari2016vnet,
  title     = {{V-Net}: Fully Convolutional Neural Networks for Volumetric Medical Image Segmentation},
  author    = {Milletari, Fausto and Navab, Nassir and Ahmadi, Seyed-Ahmad},
  booktitle = {International Conference on 3D Vision},
  pages     = {565--571},
  year      = {2016}
}

@inproceedings{murphy2025deep,
  title={Deep Learning Approaches for Cloud Property Retrieval: Comparing Foundation Model Fine-Tuning with Training From Scratch},
  author={Murphy, Danielle and Zhang, Kevin and Parten, Caleb and Sterling, Autumn and Zhang, Haoxiang and Li, Xingyan and Caraballo-Vega, Jordan A and Gong, Jie and Carroll, Mark L and Wang, Jianwu},
  booktitle={2025 IEEE International Conference on Data Mining Workshops (ICDMW)},
  pages={2816--2825},
  year={2025},
  doi={10.1109/ICDMW69685.2025.00360},
  organization={IEEE}
}

\clearpage
\appendix
\raggedbottom
\footnotesize
\setlength{\floatsep}{2pt plus 1pt minus 1pt}
\setlength{\textfloatsep}{2pt plus 1pt minus 1pt}
\setlength{\intextsep}{2pt plus 1pt minus 1pt}
\setlength{\abovecaptionskip}{1pt}
\setlength{\belowcaptionskip}{0pt}
\captionsetup[table]{font=scriptsize,skip=1pt}
\newcommand{\appmapfont}{\fontsize{4.8}{5.05}\selectfont}

\section{Spectral Interface and Band-Selection Diagnostics}
\label{app:spectral_interface_diagnostics}

This appendix lists the spectral metadata and \bandsel{} mappings behind Table~\ref{tab:spectral_transfer_diagnostics}.

\begin{table}[H]
  \centering
  \scriptsize
  \setlength{\tabcolsep}{1pt}
  \renewcommand{\arraystretch}{0.80}
  \caption{Pretrained spectral interfaces.}
  \label{tab:appendix_source_band_metadata}
  \begin{tabular}{@{}>{\raggedright\arraybackslash}p{0.25\columnwidth}p{0.08\columnwidth}>{\raggedright\arraybackslash}p{0.30\columnwidth}p{0.14\columnwidth}p{0.14\columnwidth}@{}}
\toprule
Backbone/source & Index & Band name & Center (nm) & FWHM (nm) \\
\midrule
\multirow{6}{=}{Prithvi pretraining} & 0 & BLUE & 490.0 & 65.0 \\
 & 1 & GREEN & 560.0 & 35.0 \\
 & 2 & RED & 665.0 & 30.0 \\
 & 3 & NIR\_NARROW & 865.0 & 20.0 \\
 & 4 & SWIR\_1 & 1610.0 & 90.0 \\
 & 5 & SWIR\_2 & 2202.0 & 180.0 \\
\midrule
\multirow{3}{=}{ScaleMAE pretraining} & 0 & RED & 665.0 & 80.0 \\
 & 1 & GREEN & 560.0 & 80.0 \\
 & 2 & BLUE & 490.0 & 80.0 \\
\midrule
\multirow{10}{=}{SatMAE pretraining} & 0 & B2\_BLUE & 490.0 & 65.0 \\
 & 1 & B3\_GREEN & 560.0 & 35.0 \\
 & 2 & B4\_RED & 665.0 & 30.0 \\
 & 3 & B5\_RE1 & 705.0 & 15.0 \\
 & 4 & B6\_RE2 & 740.0 & 15.0 \\
 & 5 & B7\_RE3 & 783.0 & 20.0 \\
 & 6 & B8\_NIR & 842.0 & 115.0 \\
 & 7 & B8A\_RE4 & 865.0 & 20.0 \\
 & 8 & B11\_SWIR1 & 1610.0 & 90.0 \\
 & 9 & B12\_SWIR2 & 2202.0 & 180.0 \\
\bottomrule
  \end{tabular}
\end{table}

\begin{table}[H]
  \centering
  \scriptsize
  \setlength{\tabcolsep}{1pt}
  \renewcommand{\arraystretch}{0.80}
  \caption{Downstream target-channel metadata.}
  \label{tab:appendix_target_band_metadata}
  \begin{tabular}{@{}>{\raggedright\arraybackslash}p{0.23\columnwidth}p{0.08\columnwidth}>{\raggedright\arraybackslash}p{0.34\columnwidth}p{0.13\columnwidth}p{0.13\columnwidth}@{}}
\toprule
Dataset & Index & Band name & Center (nm) & FWHM (nm) \\
\midrule
\multirow{13}{=}{Sen1Floods11} & 0 & B1\_COASTAL & 443.0 & 20.0 \\
 & 1 & B2\_BLUE & 490.0 & 65.0 \\
 & 2 & B3\_GREEN & 560.0 & 35.0 \\
 & 3 & B4\_RED & 665.0 & 30.0 \\
 & 4 & B5\_RE1 & 705.0 & 15.0 \\
 & 5 & B6\_RE2 & 740.0 & 15.0 \\
 & 6 & B7\_RE3 & 783.0 & 20.0 \\
 & 7 & B8\_NIR & 842.0 & 115.0 \\
 & 8 & B8A\_RE4 & 865.0 & 20.0 \\
 & 9 & B9\_WATER\_VAPOR & 945.0 & 20.0 \\
 & 10 & B10\_CIRRUS & 1375.0 & 30.0 \\
 & 11 & B11\_SWIR1 & 1610.0 & 90.0 \\
 & 12 & B12\_SWIR2 & 2202.0 & 180.0 \\
\midrule
\multirow{6}{=}{FireScars} & 0 & BLUE & 490.0 & 65.0 \\
 & 1 & GREEN & 560.0 & 35.0 \\
 & 2 & RED & 665.0 & 30.0 \\
 & 3 & NIR\_NARROW & 865.0 & 20.0 \\
 & 4 & SWIR\_1 & 1610.0 & 90.0 \\
 & 5 & SWIR\_2 & 2202.0 & 180.0 \\
\midrule
\multirow{14}{=}{Landslide4Sense} & 0 & B1\_COASTAL & 443.0 & 20.0 \\
 & 1 & B2\_BLUE & 490.0 & 65.0 \\
 & 2 & B3\_GREEN & 560.0 & 35.0 \\
 & 3 & B4\_RED & 665.0 & 30.0 \\
 & 4 & B5\_RE1 & 705.0 & 15.0 \\
 & 5 & B6\_RE2 & 740.0 & 15.0 \\
 & 6 & B7\_RE3 & 783.0 & 20.0 \\
 & 7 & B8\_NIR & 842.0 & 115.0 \\
 & 8 & B8A\_RE4 & 865.0 & 20.0 \\
 & 9 & B9\_WATER\_VAPOR & 945.0 & 20.0 \\
 & 10 & B10\_CIRRUS & 1375.0 & 30.0 \\
 & 11 & B11\_SWIR1 & 1610.0 & 90.0 \\
 & 12 & B12\_SWIR2 & 2202.0 & 180.0 \\
 & 13 & AUX & 0.0 & 0.0 \\
\midrule
\multirow{12}{=}{SA Crop Type} & 0 & B1\_COASTAL & 443.0 & 20.0 \\
 & 1 & B2\_BLUE & 490.0 & 65.0 \\
 & 2 & B3\_GREEN & 560.0 & 35.0 \\
 & 3 & B4\_RED & 665.0 & 30.0 \\
 & 4 & B5\_RE1 & 705.0 & 15.0 \\
 & 5 & B6\_RE2 & 740.0 & 15.0 \\
 & 6 & B7\_RE3 & 783.0 & 20.0 \\
 & 7 & B8\_NIR & 842.0 & 115.0 \\
 & 8 & B8A\_RE4 & 865.0 & 20.0 \\
 & 9 & B9\_WATER\_VAPOR & 945.0 & 20.0 \\
 & 10 & B11\_SWIR1 & 1610.0 & 90.0 \\
 & 11 & B12\_SWIR2 & 2202.0 & 180.0 \\
\bottomrule
  \end{tabular}
\end{table}

\begin{table}[H]
  \centering
  \scriptsize
  \setlength{\tabcolsep}{1pt}
  \renewcommand{\arraystretch}{0.80}
  \caption{Prithvi \bandsel{} source-to-target mapping.}
  \label{tab:appendix_band_mapping_prithvi}
  \begin{tabular}{@{}>{\raggedright\arraybackslash}p{0.18\columnwidth}p{0.10\columnwidth}>{\raggedright\arraybackslash}p{0.24\columnwidth}p{0.12\columnwidth}>{\raggedright\arraybackslash}p{0.30\columnwidth}@{}}
\toprule
Dataset & Target idx & Target band & Status & Matched native band(s) \\
\midrule
\multirow{13}{=}{Sen1Floods11} & 0 & B1\_COASTAL & extra & -- \\
 & 1 & B2\_BLUE & selected & BLUE \\
 & 2 & B3\_GREEN & selected & GREEN \\
 & 3 & B4\_RED & selected & RED \\
 & 4 & B5\_RE1 & extra & -- \\
 & 5 & B6\_RE2 & extra & -- \\
 & 6 & B7\_RE3 & extra & -- \\
 & 7 & B8\_NIR & extra & -- \\
 & 8 & B8A\_RE4 & selected & NIR\_NARROW \\
 & 9 & B9\_WATER\_VAPOR & extra & -- \\
 & 10 & B10\_CIRRUS & extra & -- \\
 & 11 & B11\_SWIR1 & selected & SWIR\_1 \\
 & 12 & B12\_SWIR2 & selected & SWIR\_2 \\
\midrule
\multirow{6}{=}{FireScars} & 0 & BLUE & selected & BLUE \\
 & 1 & GREEN & selected & GREEN \\
 & 2 & RED & selected & RED \\
 & 3 & NIR\_NARROW & selected & NIR\_NARROW \\
 & 4 & SWIR\_1 & selected & SWIR\_1 \\
 & 5 & SWIR\_2 & selected & SWIR\_2 \\
\midrule
\multirow{14}{=}{Landslide4Sense} & 0 & B1\_COASTAL & extra & -- \\
 & 1 & B2\_BLUE & selected & BLUE \\
 & 2 & B3\_GREEN & selected & GREEN \\
 & 3 & B4\_RED & selected & RED \\
 & 4 & B5\_RE1 & extra & -- \\
 & 5 & B6\_RE2 & extra & -- \\
 & 6 & B7\_RE3 & extra & -- \\
 & 7 & B8\_NIR & extra & -- \\
 & 8 & B8A\_RE4 & selected & NIR\_NARROW \\
 & 9 & B9\_WATER\_VAPOR & extra & -- \\
 & 10 & B10\_CIRRUS & extra & -- \\
 & 11 & B11\_SWIR1 & selected & SWIR\_1 \\
 & 12 & B12\_SWIR2 & selected & SWIR\_2 \\
 & 13 & AUX & auxiliary & -- \\
\midrule
\multirow{12}{=}{SA Crop Type} & 0 & B1\_COASTAL & extra & -- \\
 & 1 & B2\_BLUE & selected & BLUE \\
 & 2 & B3\_GREEN & selected & GREEN \\
 & 3 & B4\_RED & selected & RED \\
 & 4 & B5\_RE1 & extra & -- \\
 & 5 & B6\_RE2 & extra & -- \\
 & 6 & B7\_RE3 & extra & -- \\
 & 7 & B8\_NIR & extra & -- \\
 & 8 & B8A\_RE4 & selected & NIR\_NARROW \\
 & 9 & B9\_WATER\_VAPOR & extra & -- \\
 & 10 & B11\_SWIR1 & selected & SWIR\_1 \\
 & 11 & B12\_SWIR2 & selected & SWIR\_2 \\
\bottomrule
  \end{tabular}
\end{table}

\begin{table}[H]
  \centering
  \scriptsize
  \setlength{\tabcolsep}{1pt}
  \renewcommand{\arraystretch}{0.8}
  \caption{ScaleMAE \bandsel{} source-to-target mapping.}
  \label{tab:appendix_band_mapping_scalemae}
  \begin{tabular}{@{}>{\raggedright\arraybackslash}p{0.18\columnwidth}p{0.10\columnwidth}>{\raggedright\arraybackslash}p{0.24\columnwidth}p{0.12\columnwidth}>{\raggedright\arraybackslash}p{0.30\columnwidth}@{}}
\toprule
Dataset & Target idx & Target band & Status & Matched native band(s) \\
\midrule
\multirow{13}{=}{Sen1Floods11} & 0 & B1\_COASTAL & extra & -- \\
 & 1 & B2\_BLUE & selected & BLUE \\
 & 2 & B3\_GREEN & selected & GREEN \\
 & 3 & B4\_RED & selected & RED \\
 & 4 & B5\_RE1 & extra & -- \\
 & 5 & B6\_RE2 & extra & -- \\
 & 6 & B7\_RE3 & extra & -- \\
 & 7 & B8\_NIR & extra & -- \\
 & 8 & B8A\_RE4 & extra & -- \\
 & 9 & B9\_WATER\_VAPOR & extra & -- \\
 & 10 & B10\_CIRRUS & extra & -- \\
 & 11 & B11\_SWIR1 & extra & -- \\
 & 12 & B12\_SWIR2 & extra & -- \\
\midrule
\multirow{6}{=}{FireScars} & 0 & BLUE & selected & BLUE \\
 & 1 & GREEN & selected & GREEN \\
 & 2 & RED & selected & RED \\
 & 3 & NIR\_NARROW & extra & -- \\
 & 4 & SWIR\_1 & extra & -- \\
 & 5 & SWIR\_2 & extra & -- \\
\midrule
\multirow{14}{=}{Landslide4Sense} & 0 & B1\_COASTAL & extra & -- \\
 & 1 & B2\_BLUE & selected & BLUE \\
 & 2 & B3\_GREEN & selected & GREEN \\
 & 3 & B4\_RED & selected & RED \\
 & 4 & B5\_RE1 & extra & -- \\
 & 5 & B6\_RE2 & extra & -- \\
 & 6 & B7\_RE3 & extra & -- \\
 & 7 & B8\_NIR & extra & -- \\
 & 8 & B8A\_RE4 & extra & -- \\
 & 9 & B9\_WATER\_VAPOR & extra & -- \\
 & 10 & B10\_CIRRUS & extra & -- \\
 & 11 & B11\_SWIR1 & extra & -- \\
 & 12 & B12\_SWIR2 & extra & -- \\
 & 13 & AUX & auxiliary & -- \\
\midrule
\multirow{12}{=}{SA Crop Type} & 0 & B1\_COASTAL & extra & -- \\
 & 1 & B2\_BLUE & selected & BLUE \\
 & 2 & B3\_GREEN & selected & GREEN \\
 & 3 & B4\_RED & selected & RED \\
 & 4 & B5\_RE1 & extra & -- \\
 & 5 & B6\_RE2 & extra & -- \\
 & 6 & B7\_RE3 & extra & -- \\
 & 7 & B8\_NIR & extra & -- \\
 & 8 & B8A\_RE4 & extra & -- \\
 & 9 & B9\_WATER\_VAPOR & extra & -- \\
 & 10 & B11\_SWIR1 & extra & -- \\
 & 11 & B12\_SWIR2 & extra & -- \\
\bottomrule
  \end{tabular}
\end{table}

\begin{table}[H]
  \centering
  \scriptsize
  \setlength{\tabcolsep}{1pt}
  \renewcommand{\arraystretch}{0.80}
  \caption{SatMAE \bandsel{} source-to-target mapping.}
  \label{tab:appendix_band_mapping_satmae}
  \begin{tabular}{@{}>{\raggedright\arraybackslash}p{0.18\columnwidth}p{0.10\columnwidth}>{\raggedright\arraybackslash}p{0.24\columnwidth}p{0.12\columnwidth}>{\raggedright\arraybackslash}p{0.30\columnwidth}@{}}
\toprule
Dataset & Target idx & Target band & Status & Matched native band(s) \\
\midrule
\multirow{13}{=}{Sen1Floods11} & 0 & B1\_COASTAL & extra & -- \\
 & 1 & B2\_BLUE & selected & B2\_BLUE \\
 & 2 & B3\_GREEN & selected & B3\_GREEN \\
 & 3 & B4\_RED & selected & B4\_RED \\
 & 4 & B5\_RE1 & selected & B5\_RE1 \\
 & 5 & B6\_RE2 & selected & B6\_RE2 \\
 & 6 & B7\_RE3 & selected & B7\_RE3 \\
 & 7 & B8\_NIR & selected & B8\_NIR \\
 & 8 & B8A\_RE4 & selected & B8A\_RE4 \\
 & 9 & B9\_WATER\_VAPOR & extra & -- \\
 & 10 & B10\_CIRRUS & extra & -- \\
 & 11 & B11\_SWIR1 & selected & B11\_SWIR1 \\
 & 12 & B12\_SWIR2 & selected & B12\_SWIR2 \\
\midrule
\multirow{6}{=}{FireScars} & 0 & BLUE & selected & B2\_BLUE \\
 & 1 & GREEN & selected & B3\_GREEN \\
 & 2 & RED & selected & B4\_RED; B5\_RE1; B6\_RE2 \\
 & 3 & NIR\_NARROW & selected & B7\_RE3; B8\_NIR; B8A\_RE4 \\
 & 4 & SWIR\_1 & selected & B11\_SWIR1 \\
 & 5 & SWIR\_2 & selected & B12\_SWIR2 \\
\midrule
\multirow{14}{=}{Landslide4Sense} & 0 & B1\_COASTAL & extra & -- \\
 & 1 & B2\_BLUE & selected & B2\_BLUE \\
 & 2 & B3\_GREEN & selected & B3\_GREEN \\
 & 3 & B4\_RED & selected & B4\_RED \\
 & 4 & B5\_RE1 & selected & B5\_RE1 \\
 & 5 & B6\_RE2 & selected & B6\_RE2 \\
 & 6 & B7\_RE3 & selected & B7\_RE3 \\
 & 7 & B8\_NIR & selected & B8\_NIR \\
 & 8 & B8A\_RE4 & selected & B8A\_RE4 \\
 & 9 & B9\_WATER\_VAPOR & extra & -- \\
 & 10 & B10\_CIRRUS & extra & -- \\
 & 11 & B11\_SWIR1 & selected & B11\_SWIR1 \\
 & 12 & B12\_SWIR2 & selected & B12\_SWIR2 \\
 & 13 & AUX & auxiliary & -- \\
\midrule
\multirow{12}{=}{SA Crop Type} & 0 & B1\_COASTAL & extra & -- \\
 & 1 & B2\_BLUE & selected & B2\_BLUE \\
 & 2 & B3\_GREEN & selected & B3\_GREEN \\
 & 3 & B4\_RED & selected & B4\_RED \\
 & 4 & B5\_RE1 & selected & B5\_RE1 \\
 & 5 & B6\_RE2 & selected & B6\_RE2 \\
 & 6 & B7\_RE3 & selected & B7\_RE3 \\
 & 7 & B8\_NIR & selected & B8\_NIR \\
 & 8 & B8A\_RE4 & selected & B8A\_RE4 \\
 & 9 & B9\_WATER\_VAPOR & extra & -- \\
 & 10 & B11\_SWIR1 & selected & B11\_SWIR1 \\
 & 11 & B12\_SWIR2 & selected & B12\_SWIR2 \\
\bottomrule
  \end{tabular}
\end{table}

\section{Full Direct-MLP Companion Results}
\label{app:full_mlp_results}

Tables~\ref{tab:appendix_full_mlp_prithvi}--\ref{tab:appendix_full_mlp_satmae}
expand the direct-MLP companion experiments summarized in
Table~\ref{tab:mlp_bre_adapter_summary}. Each row keeps the GeoFM, dataset, and
baseline fine-tuning policy fixed, and changes only the input adapter from
\bandsel{} to a direct all-band MLP projection. These results do not include
\starplanner{}. Values are seed mean ${\pm}$ sample standard deviation test
macro \miou{} (\%) over seeds 42--44; $\Delta$ is +MLP minus the matched
\bandsel{} baseline.

\begin{table}[H]
  \centering
  \scriptsize
  \setlength{\tabcolsep}{1.3pt}
  \renewcommand{\arraystretch}{0.90}
  \caption{Direct-MLP companions for Prithvi.}
  \label{tab:appendix_full_mlp_prithvi}
  \begin{tabular}{@{}p{0.28\columnwidth}p{0.17\columnwidth}ccc@{}}
    \toprule
    Dataset & Policy & \bandsel{} & +MLP & $\Delta$ \\
    \midrule
    Sen1Floods11 & LoRA-16 & $84.46{\pm}1.91$ & $84.24{\pm}2.00$ & $-0.22$ \\
    Sen1Floods11 & LoRA-32 & $85.10{\pm}1.70$ & $83.53{\pm}2.11$ & $-1.57$ \\
    Sen1Floods11 & LoRA-64 & $84.40{\pm}3.18$ & $84.14{\pm}2.26$ & $-0.26$ \\
    Sen1Floods11 & Surgical & $84.81{\pm}0.27$ & $83.40{\pm}1.99$ & $-1.41$ \\
    Sen1Floods11 & Full-FT & $86.24{\pm}2.95$ & $86.25{\pm}2.12$ & $+0.00$ \\
    FireScars & LoRA-16 & $85.25{\pm}0.35$ & $85.05{\pm}0.68$ & $-0.20$ \\
    FireScars & LoRA-32 & $85.47{\pm}0.35$ & $85.03{\pm}0.78$ & $-0.43$ \\
    FireScars & LoRA-64 & $85.56{\pm}0.45$ & $85.78{\pm}1.46$ & $+0.21$ \\
    FireScars & Surgical & $85.52{\pm}0.58$ & $83.37{\pm}1.71$ & $-2.16$ \\
    FireScars & Full-FT & $87.71{\pm}1.13$ & $85.31{\pm}0.72$ & $-2.40$ \\
    Landslide4Sense & LoRA-16 & $76.88{\pm}0.39$ & $76.16{\pm}0.51$ & $-0.73$ \\
    Landslide4Sense & LoRA-32 & $77.07{\pm}0.05$ & $75.88{\pm}1.31$ & $-1.20$ \\
    Landslide4Sense & LoRA-64 & $77.00{\pm}0.30$ & $76.30{\pm}0.36$ & $-0.70$ \\
    Landslide4Sense & Surgical & $76.32{\pm}0.17$ & $75.91{\pm}0.05$ & $-0.41$ \\
    Landslide4Sense & Full-FT & $77.25{\pm}0.08$ & $76.37{\pm}0.50$ & $-0.88$ \\
    SA Crop Type & LoRA-16 & $35.46{\pm}0.53$ & $32.57{\pm}0.17$ & $-2.89$ \\
    SA Crop Type & LoRA-32 & $35.61{\pm}0.56$ & $33.18{\pm}0.40$ & $-2.43$ \\
    SA Crop Type & LoRA-64 & $35.87{\pm}0.14$ & $33.09{\pm}0.27$ & $-2.78$ \\
    SA Crop Type & Surgical & $34.08{\pm}0.44$ & $30.98{\pm}0.41$ & $-3.10$ \\
    SA Crop Type & Full-FT & $37.63{\pm}1.01$ & $34.67{\pm}0.23$ & $-2.97$ \\
    \bottomrule
  \end{tabular}
\end{table}

\begin{table}[H]
  \centering
  \scriptsize
  \setlength{\tabcolsep}{1.3pt}
  \renewcommand{\arraystretch}{0.90}
  \caption{Direct-MLP companions for ScaleMAE.}
  \label{tab:appendix_full_mlp_scalemae}
  \begin{tabular}{@{}p{0.28\columnwidth}p{0.17\columnwidth}ccc@{}}
    \toprule
    Dataset & Policy & \bandsel{} & +MLP & $\Delta$ \\
    \midrule
    Sen1Floods11 & LoRA-16 & $76.27{\pm}2.71$ & $81.52{\pm}2.38$ & $+5.25$ \\
    Sen1Floods11 & LoRA-32 & $77.07{\pm}2.51$ & $79.84{\pm}2.35$ & $+2.77$ \\
    Sen1Floods11 & LoRA-64 & $77.17{\pm}2.16$ & $79.92{\pm}3.96$ & $+2.75$ \\
    Sen1Floods11 & Surgical & $73.39{\pm}3.01$ & $79.92{\pm}0.58$ & $+6.54$ \\
    Sen1Floods11 & Full-FT & $75.41{\pm}1.22$ & $83.76{\pm}1.76$ & $+8.35$ \\
    FireScars & LoRA-16 & $67.86{\pm}1.96$ & $64.29{\pm}4.60$ & $-3.57$ \\
    FireScars & LoRA-32 & $68.20{\pm}1.89$ & $55.30{\pm}20.09$ & $-12.90$ \\
    FireScars & LoRA-64 & $69.03{\pm}0.91$ & $57.60{\pm}21.83$ & $-11.43$ \\
    FireScars & Surgical & $68.27{\pm}1.56$ & $60.72{\pm}7.67$ & $-7.55$ \\
    FireScars & Full-FT & $72.30{\pm}1.39$ & $67.97{\pm}16.76$ & $-4.33$ \\
    Landslide4Sense & LoRA-16 & $75.05{\pm}0.20$ & $73.30{\pm}2.96$ & $-1.75$ \\
    Landslide4Sense & LoRA-32 & $75.16{\pm}0.28$ & $72.63{\pm}2.77$ & $-2.54$ \\
    Landslide4Sense & LoRA-64 & $75.15{\pm}0.19$ & $72.63{\pm}4.12$ & $-2.52$ \\
    Landslide4Sense & Surgical & $74.82{\pm}0.31$ & $68.82{\pm}9.41$ & $-6.00$ \\
    Landslide4Sense & Full-FT & $75.36{\pm}0.07$ & $74.16{\pm}0.60$ & $-1.20$ \\
    SA Crop Type & LoRA-16 & $31.50{\pm}0.44$ & $30.92{\pm}1.09$ & $-0.58$ \\
    SA Crop Type & LoRA-32 & $31.90{\pm}0.34$ & $30.64{\pm}1.40$ & $-1.27$ \\
    SA Crop Type & LoRA-64 & $32.09{\pm}0.35$ & $30.46{\pm}1.94$ & $-1.62$ \\
    SA Crop Type & Surgical & $31.03{\pm}0.33$ & $30.50{\pm}1.53$ & $-0.53$ \\
    SA Crop Type & Full-FT & $33.22{\pm}0.52$ & $27.63{\pm}7.95$ & $-5.59$ \\
    \bottomrule
  \end{tabular}
\end{table}

\begin{table}[H]
  \centering
  \scriptsize
  \setlength{\tabcolsep}{1.3pt}
  \renewcommand{\arraystretch}{0.90}
  \caption{Direct-MLP companions for SatMAE.}
  \label{tab:appendix_full_mlp_satmae}
  \begin{tabular}{@{}p{0.28\columnwidth}p{0.17\columnwidth}ccc@{}}
    \toprule
    Dataset & Policy & \bandsel{} & +MLP & $\Delta$ \\
    \midrule
    Sen1Floods11 & LoRA-16 & $83.76{\pm}1.45$ & $83.48{\pm}1.30$ & $-0.28$ \\
    Sen1Floods11 & LoRA-32 & $84.95{\pm}1.82$ & $84.07{\pm}0.43$ & $-0.87$ \\
    Sen1Floods11 & LoRA-64 & $84.07{\pm}1.41$ & $84.81{\pm}0.12$ & $+0.74$ \\
    Sen1Floods11 & Surgical & $82.42{\pm}3.31$ & $83.05{\pm}2.79$ & $+0.63$ \\
    Sen1Floods11 & Full-FT & $85.94{\pm}2.08$ & $85.22{\pm}0.56$ & $-0.73$ \\
    FireScars & LoRA-16 & $64.30{\pm}0.68$ & $58.31{\pm}4.75$ & $-5.99$ \\
    FireScars & LoRA-32 & $64.97{\pm}0.30$ & $61.08{\pm}5.41$ & $-3.88$ \\
    FireScars & LoRA-64 & $66.79{\pm}0.63$ & $59.98{\pm}7.61$ & $-6.81$ \\
    FireScars & Surgical & $65.15{\pm}2.02$ & $59.67{\pm}7.30$ & $-5.48$ \\
    FireScars & Full-FT & $71.61{\pm}0.71$ & $59.00{\pm}4.19$ & $-12.61$ \\
    Landslide4Sense & LoRA-16 & $77.46{\pm}0.04$ & $75.43{\pm}0.79$ & $-2.04$ \\
    Landslide4Sense & LoRA-32 & $77.56{\pm}0.16$ & $76.08{\pm}0.86$ & $-1.48$ \\
    Landslide4Sense & LoRA-64 & $77.58{\pm}0.13$ & $76.00{\pm}0.19$ & $-1.58$ \\
    Landslide4Sense & Surgical & $77.37{\pm}0.25$ & $75.19{\pm}0.75$ & $-2.19$ \\
    Landslide4Sense & Full-FT & $77.26{\pm}0.13$ & $75.46{\pm}0.91$ & $-1.80$ \\
    SA Crop Type & LoRA-16 & $32.42{\pm}0.35$ & $25.89{\pm}1.12$ & $-6.53$ \\
    SA Crop Type & LoRA-32 & $32.77{\pm}0.41$ & $25.82{\pm}0.90$ & $-6.95$ \\
    SA Crop Type & LoRA-64 & $32.69{\pm}0.14$ & $25.33{\pm}3.95$ & $-7.35$ \\
    SA Crop Type & Surgical & $31.20{\pm}0.63$ & $26.88{\pm}0.20$ & $-4.32$ \\
    SA Crop Type & Full-FT & $33.30{\pm}1.12$ & $30.12{\pm}0.20$ & $-3.18$ \\
    \bottomrule
  \end{tabular}
\end{table}

\section{Comparison with Published Task-Specific Fine-Tuning Results}
  \label{app:published_prithvi_comparison}

Table~\ref{tab:appendix_published_prithvi_comparison} compares our results with published Prithvi-EO-2.0 task-specific fine-tuning results~\cite{roy2024prithvi}. These numbers should be interpreted as practical context rather than a controlled apples-to-apples benchmark, because the published results use task-specific training recipes, model variants, losses, and repeated hyperparameter search, while our experiments use a controlled shared segmentation setup across all experiments to compare cross-sensor adaptation methods.

  \begin{table}[th]
    \centering
    \scriptsize
    \setlength{\tabcolsep}{3pt}
    \renewcommand{\arraystretch}{0.95}
    \caption{Comparison with published Prithvi-EO-2.0 task-specific fine-tuning results. Published values are from Prithvi-EO-
    2.0~\cite{roy2024prithvi}; our values are from Table~\ref{tab:main_comparison}. The evaluation metric in the table is \miou{}.}
    \label{tab:appendix_published_prithvi_comparison}
    \begin{tabular}{@{}p{0.08\textwidth}p{0.12\textwidth}p{0.12\textwidth}p{0.13\textwidth}@{}}
      \toprule
      Dataset & Published result & Our Prithvi+\method{} & Best \method{} in this paper \\
      \midrule
      Sen1Floods11 & $90.3$ & $86.39{\pm}0.85$ & $87.19{\pm}0.82$ 
      (SatMAE) \\
      FireScars & $90.5$  & $85.40{\pm}0.33$ & $85.40{\pm}0.33$
      (Prithvi) \\
      Landslide4Sense &  $70.4$  & $77.28{\pm}0.09$  & $78.26{\pm}0.13$ 
      (SatMAE) \\
      SA Crop Type & $41.70{\pm}0.26$ & $36.32{\pm}0.63$ & $36.32{\pm}0.63$ 
      (Prithvi) \\
      \bottomrule
    \end{tabular}
  \end{table}

\end{document}